\documentclass[10pt,twocolumn,letterpaper]{article}

\usepackage[pagenumbers]{cvpr} % To force page numbers, e.g. for an arXiv version

\usepackage{kotex}
\usepackage{multirow}
\usepackage{colortbl}
\usepackage{xcolor}
\usepackage{color}

\usepackage{booktabs}
\usepackage{bbding}
\usepackage{amsmath}
\usepackage{amssymb}
\usepackage{xparse}
\usepackage{makecell}

\definecolor{orange}{HTML}{F6E3CE}

\newcommand\ro[1]{{\color{magenta}#1}}

\definecolor{cvprblue}{rgb}{0.21,0.49,0.74}
\usepackage[pagebackref,breaklinks,colorlinks,allcolors=cvprblue]{hyperref}

\def\confName{CVPR}
\def\confYear{2025}

\title{Implicit Grid Convolution for Multi-Scale Image Super-Resolution}

\author{Dongheon Lee \hspace{1cm} Seokju Yun \hspace{1cm} Youngmin Ro\thanks{Corresponding Author}\\
University of Seoul\\
\tt{Code: \ro{https://github.com/dslisleedh/IGConv}}\\
{\tt\small \{dslisleedh, wsz871, youngmin.ro\}@uos.ac.kr}
}

\begin{document}
\maketitle
\vspace{-0.6cm}
\begin{abstract}
For Image Super-Resolution~(SR), it is common to train and evaluate scale-specific models composed of an encoder and upsampler for each targeted scale. 
Consequently, many SR studies encounter substantial training times and complex deployment requirements.
In this paper, we address this limitation by training and evaluating multiple scales simultaneously. 
Notably, we observe that encoder features are similar across scales and that the Sub-Pixel Convolution~(SPConv), widely-used scale-specific upsampler, exhibits strong inter-scale correlations in its functionality.
Building on these insights, we propose a multi-scale framework that employs a single encoder in conjunction with Implicit Grid Convolution~(IGConv), our novel upsampler, which unifies SPConv across all scales within a single module.
Extensive experiments demonstrate that our framework achieves comparable performance to existing fixed-scale methods while reducing the training budget and stored parameters three-fold and maintaining the same latency. 
Additionally, we propose IGConv$^{+}$ to improve performance further by addressing spectral bias and allowing input-dependent upsampling and ensembled prediction. 
As a result, ATD-IGConv$^{+}$ achieves a notable 0.21dB improvement in PSNR on Urban100$\times$4, while also reducing the training budget, stored parameters, and inference cost compared to the existing ATD.
\end{abstract}
   
\vspace{-0.8cm}
\section{Introduction}
\vspace{-0.2cm}

\begin{figure}[t]
  \centering
  \includegraphics[width=\columnwidth]{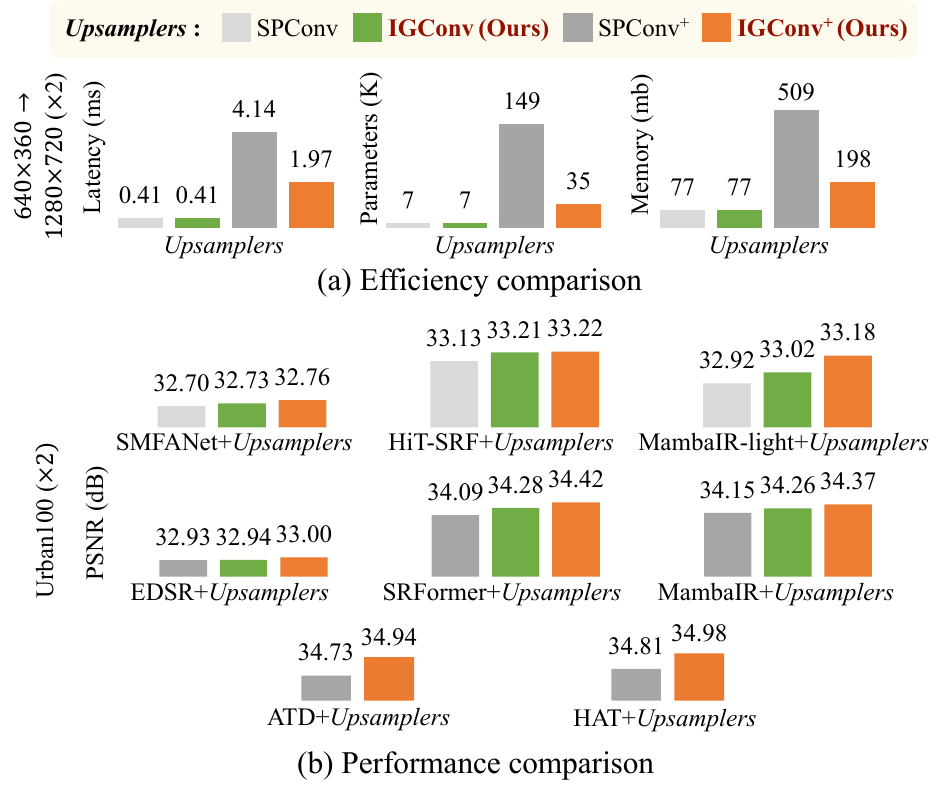}
  \vspace{-0.9cm}
  \caption{
    Efficiency and performance comparison on existing upsampler~(SPConv and SPConv$^{+}$) with our proposals~(IGConv and IGConv$^{+}$) on various metrics and models. Efficiency metrics are measured by reconstructing an HD~(1280$\times$720) image on an A6000 GPU after instantiating our proposals on a $\times2$ scale. 
  }
  \label{fig:teaser}
  \vspace{-0.8cm}
\end{figure}

Image Super-Resolution~(SR) aims to restore a High-Resolution Image~($I^{HR}$) from a Low-Resolution Image~($I^{LR}$) input, which is one of the most fundamental challenges in computer vision and graphics.
Over a decade ago, SRCNN~\cite{SRCNN} successfully introduced neural networks to SR, leading to significant performance improvements. 
Following SRCNN, many previous studies have focused on improving performance by proposing new core operators or larger models, leading to massive models like HAT-Large~\cite{HAT} that leverage up to 41 million parameters.

In general, classic SR methods train and evaluate a single scale-specific model for each target scale~\cite{SRCNN, SwinIR}. Since SR tasks typically consider three scales~($\times$2, $\times3$, and $\times$4), the training budget and stored parameters are significantly increased by a factor of three.
As the size of models and datasets increases and training strategies become more complex, the time required for training has grown substantially.
For instance, training a large model with 20 million parameters~\cite{MambaIR} for $\times$2 scale using four A6000 GPUs takes approximately 241 hours.
This issue is expected to become more pronounced in the future. 
Also, from a deployment perspective, storing and loading SR models for every target size is a significant restriction when computing resource-constrained scenarios like real-world applications.

In this paper, we present a novel multi-scale framework, developed through an in-depth investigation, that employs a single encoder and a single upsampler pair.
\begin{table}[ht]
\caption{
    Comparisons of various upsamplers employing RDN~\cite{RDN} encoder. 
    The efficiency metrics are measured by upsampling a 256$\times$256 image for scale $\times$4 using an A6000 GPU.
}\label{tab:compre_type}
\vspace{-0.4cm}
\resizebox{\columnwidth}{!}{%
\begin{tabular}{l|c|cc|ccc}
\multirow{2}{*}{Upsampler~@RDN~\cite{RDN}} & \multirow{2}{*}{Type} & \multirow{2}{*}{\begin{tabular}[c]{@{}c@{}}Latency\\ (ms)\end{tabular}} & \multirow{2}{*}{\begin{tabular}[c]{@{}c@{}}Memory\\ (mb)\end{tabular}} & \multicolumn{3}{c}{Urban100~(PSNR$\uparrow$)} \\ \cline{5-7} 
 &  &  &  & $\times$2 & $\times$3 & $\times$4 \\ \hline
SPConv$^{+}$ & Fixed & 5.5 & 529.1 & 32.89 & 28.80 & 26.61 \\
LM-LTE~\cite{LMF} & Arb. & 95.7 & 1442.4 & 33.03 & 28.96 & 26.80 \\
\textbf{IGConv$^{+}$ (Ours)} & Multi. & 3.9 & 193.5 & 33.17 & 29.11 & 26.96 \\ \hline
\end{tabular}
}
\vspace{-0.6cm}
\end{table}
In our preliminary study assessing the similarity between features extracted from different encoders trained at various scales, we observe that features from the later stages of these encoders exhibit significant similarity. 
This characteristic is consistently present across models utilizing various core operators such as convolution~\cite{SMFANet}, self-attention~\cite{HiTSR}, and state-space models~(SSM)~\cite{MambaIR}.
These findings provide valuable insight into the potential for training multiple scales with a single encoder.
Building upon this, utilizing upsamplers capable of inferring any scale, as proposed in the Arbitrary-Scale Super-Resolution (ASSR) domain, appears to be straightforward.
However, as shown in Table~\ref{tab:compre_type}, the upsamplers from ASSR require excessive computational cost for inefficient architecture to predict non-integer scales.
Therefore, we focus on the structural mechanism of the widely used scale-specific upsampler, Sub-Pixel Convolution~(SPConv), and observe that the goal of upsampling filters at different scales is highly analogous.
For example, as shown in Figure~\ref{fig:pixel_shuffle}, the filtered sub-pixels exhibit significant correlations across scales within the 2D space.
Building on this insight, we propose Implicit Grid Convolution~(IGConv), which unifies SPConv at all scales, enabling multi-scale predictions while maintaining the same latency.

Moreover, we propose IGConv$^{+}$, which boosts performance further by addressing spectral bias and enabling input-dependent upsampling and ensembled prediction. 
We leverage frequency loss to mitigate spectral bias and introduce Implicit Grid Sampling~(IGSample) designed to handle both spectral bias and input-dependent upsampling.
Additionally, we introduce a feature-level geometric re-parameterization~(FGRep), which enables ensemble prediction with a single forward pass.
As a result, applying IGConv to existing scales-specific methods achieves comparable performance while reducing the training budget and stored parameters by one-third and maintaining the same latency.
Furthermore, applying IGConv$^{+}$ to methods such as EDSR~\cite{EDSR}, SRFormer~\cite{SRFormer}, and MambaIR~\cite{MambaIR} improves PSNR by 0.16 dB, 0.25 dB, and 0.12 dB, respectively, on Urban100$\times$4 still reducing substantial training overheads.
Moreover, even for large-size models adopting ImageNet pre-training strategy~\cite{HAT}, our methods impressively reduce training time by up to 552 hours.
% it achieves an impressive reduction in training time of up to 552 hours.

Our contributions are summarized as follows:
\begin{itemize}
    \item We highlight the inefficiency of the classic fixed-scale approaches and address it by proposing the multi-scale frame employing a single encoder and IGConv.
    \item Furthermore, we propose IGConv$^{+}$, which improves performance by employing frequency loss and introducing IGSample, and FGRep.
    \item As a result, SRFormer-IGConv$^{+}$ achieves remarkable 0.33 dB improvement on Urban100$\times$2 compared to the existing SRFormer, as shown in Figure~\ref{fig:teaser}.
\end{itemize}

\section{Related Work}
\vspace{-0.2cm}
% \begin{table}[t]
%     \begin{minipage}{0.5\textwidth}
%     \centering
%         \vspace{-4.1mm}
%         \tablestyle{3.2pt}{0.95}
%         \input{tab/smr}
%         \vspace{-2.2mm}
%         \caption{\textbf{Singular Modulation Ratio Comparison.} }\label{tab: smr}
%         \vspace{-6mm}
%     \end{minipage}
% \end{table}

% \vspace{-0.15cm}
% \subsection{Classic Image Super-Resolution}
% \vspace{-0.15cm}

\noindent\textbf{Classic Image Super-Resolution} From early on to the present, CNN-based methods, which primarily utilize convolution operations suited for image processing due to their local bias and translation invariance, have been foundational in SR tasks~\cite{SRCNN, ESPCN, EDSR, RCAN, SMFANet}.
Recently, transformers~\cite{Transformers, ViT} have garnered significant attention in SR tasks due to their ability to handle long-range dependencies and their advantage of leveraging dynamic weights.  
Methods that compute self-attention within a window patch~\cite{SwinIR} or in a transposed manner~(channel-wise)~\cite{Restormer} to reduce the number of pixels processed at once for the quadratic complexity of self-attention have demonstrated superior performance with reduced computational complexity and parameters.
Building on the success of window/transposed self-attention, studies have continued to report improvements in various aspects: widening receptive fields~\cite{HAT, SRFormer, CFAT, ATD}, spectral bias~\cite{CRAFT}, quadratic complexity~\cite{RGT, HiTSR}, and memory inefficiency~\cite{ELAN, DITN}. 
In contrast, MambaIR~\cite{MambaIR} successfully introduced SSM~\cite{Mamba}, a promising alternative to self-attention, to low-level vision tasks including SR by enhancing their local mixing ability and intermediate feature representation.

Almost all listed studies adopt the fixed-scale approach employing SPConv for upsampling.
Instead, we propose a multi-scale framework employing a single encoder and IGConv to train multiple scales simultaneously.

% \vspace{-0.15cm}
% \subsection{Multi/Arbitrary-Scale Image Super-Resolution}
% \vspace{-0.15cm}
\noindent\textbf{Multi/Arbitrary-Scale Super-Resolution}
Dissimilar from the classic image SR methods, there have been experimental studies that can predict more than a single scale.
For example, LapSRN~\cite{LapSRN} proposed progressively upsampling architecture to predict $\times$8 scale reliably, MDSR~\cite{EDSR} shared a feature extractor across three scales with scale-specific heads and tails.
MetaSR~\cite{MetaSR} proposed the meta-upscaling module that performs convolution with pixel-wise dynamic filters for ASSR.
Recently, research in the ASSR field has gained significant attention by adopting Implicit Neural Representation~(INR) from the graphics domain~\cite{NERF}. 
LIIF~\cite{LIIF} predicts RGB employing MLPs with 2D relative position, nearby four feature vectors, and cell decoding.
Subsequent studies have focused on improving aspects such as the spectral bias~\cite{LTE}, local ensemble~\cite{CiaoSR, CLIT}, scale-equivalence~\cite{EQSR}, and efficiency~\cite{OPESR, LMF, CUF}.

While our approach shares similarities with ASSR methods by training multiple scales simultaneously using INR-based methods, it differs in that we do not specifically target arbitrary scales.
Furthermore, we demonstrate that our method is superior to existing multi-scale SR methods in Appendix~\ref{sec:multiscale_comparison}. 

\begin{figure*}[h]
  \centering
  \includegraphics[width=\textwidth]{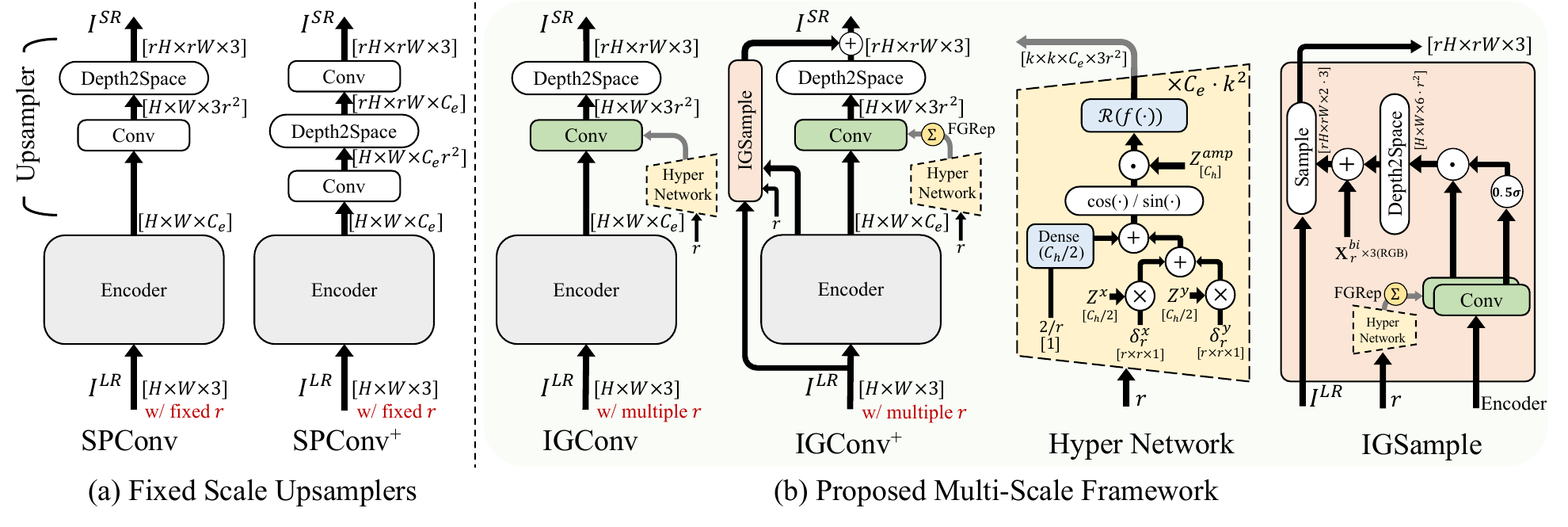}
  \vspace{-0.4cm}
  \caption{
    The structure of SR models.
    (a) illustrates the classic fixed-scale SR methods employing SPConv and SPConv~$^{+}$, while (b) illustrates our multi-scale frameworks employing IGConv, and IGConv$^{+}$.
    Our proposed methods comprise the hyper-network to generate convolution filters based on scale and employ the IGSample as a sub-module for efficient input-dependent upsampling. 
    FGRep is employed to improve performance by performing ensemble prediction with a single forward pass. 
  }
  \vspace{-0.4cm}
  \label{fig:igconvvis}
\end{figure*}

\section{Proposed Methods}
This section describes the structure of classical SR models and presents the preliminary analyses that lead us to train multiple integer scales simultaneously with a single model. 
Based on the analyses, we use a single encoder for all scales and introduce our novel upsampler, IGConv, which efficiently predicts multiple integer scales.
Following that, we describe the methods added to IGConv$^{+}$ -- frequency loss, IGSample, and FGRep -- to enhance performance further.

\vspace{-0.1cm}
\subsection{Structure of SR Models}
\vspace{-0.1cm}
As shown in Figure~\ref{fig:igconvvis}, the structure of classical SR models can be presented as:
\vspace{-0.3cm}

\begin{equation}\label{eqn:OverallArchitecture}
\begin{split}
   M = \mathcal{E}(I^{LR}), \\ %,   \textbf{F}\in \mathbb{R}^{H\times W\times C_{e}} \\ 
   I^{SR} = \mathcal{U}(M, r), % ,    I^{SR}\in \mathbb{R}^{rH\times rW\times 3}
\end{split}
\end{equation}

\noindent where $\mathcal{E}$ denotes the encoder that extracts deep feature representation $M$ $\in \mathbb{R}^{H \times W \times C_e}$ with the same resolution as the input $I^{LR}$ $\in \mathbb{R}^{H \times W \times 3}$, and $\mathcal{U}$ represents the upsampler that produces high-resolution output $I^{SR}$ $\in \mathbb{R}^{rH \times rW \times 3}$ according to a scale factor $\mathit{r}$.

\begin{figure}[ht!]
  \centering
  \includegraphics[width=\columnwidth]{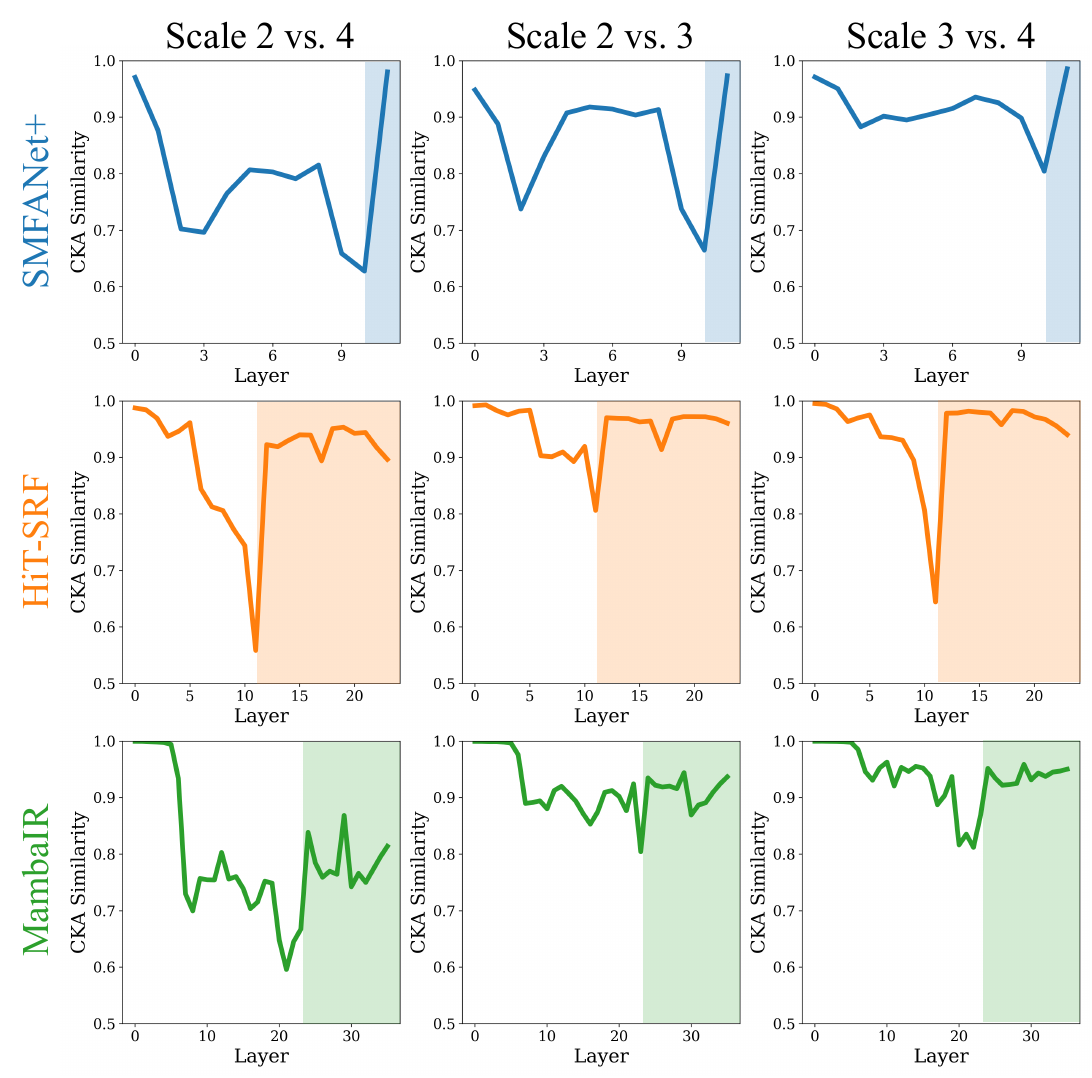}
  \caption{
    Visualization of CKA similarity~\cite{CKA} between feature maps at scale $\times$2, $\times$3, and $\times$4 varying layers of SMFANet$+$~\cite{SMFANet}, HiT-SRF~\cite{HiTSR}, and MambaIR~\cite{MambaIR}. 
    CKA similarity demonstrates that feature maps at different scales become increasingly similar as they approach the later layer.
  }
  \vspace{-0.7cm}
  \label{fig:cka}
\end{figure}

\vspace{-0.1cm}
\subsection{Preliminary Analysis}
\vspace{-0.1cm}
Numerous SR methods~\cite{EDSR, HiTSR, MambaIR, SMFANet, SRFormer, ATD, HAT} train and evaluate scale-specific models on each targeted scale, even though their encoders share the exact same structure.
The scale-specific encoders require their own training budget and storage space, significantly increasing computing resources.
This raises the question of whether the benefits of scale-specific encoders justify the significant additional computational resources required.
To verify this question, we compare the features extracted by encoders at different scales by analyzing their CKA similarity~\cite{CKA} across various core operators and model sizes~\cite{SMFANet, HiTSR, MambaIR}.ㅁ
As shown in Figure~\ref{fig:cka}, surprisingly, the CKA similarity of feature maps between different scales exceeds 0.9 on average, indicating that encoders at different scales tend to extract similar features.
This high similarity suggests that scale-specific encoders may not be necessary, given the substantial overlap in the features they capture.
Consequently, we leverage only a single encoder to train multiple integer scales.

\begin{figure}[ht]
  \centering
  \includegraphics[width=\columnwidth]{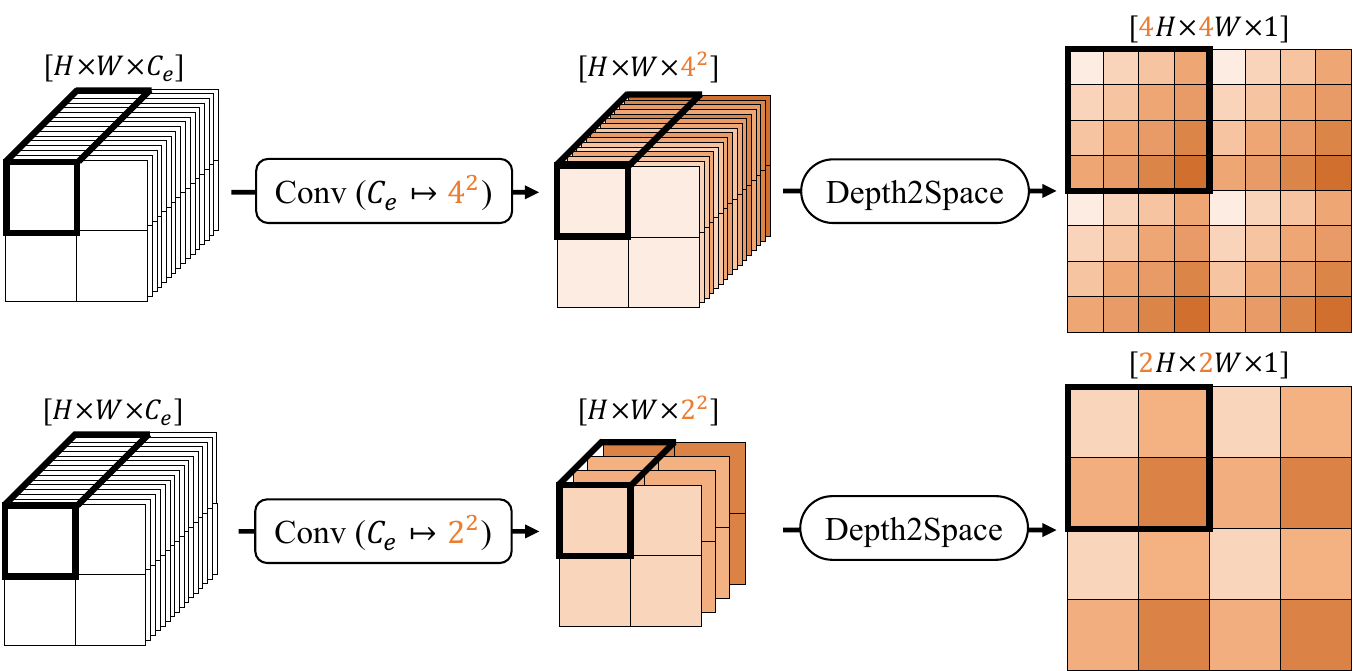}
  \caption{
    Visualization of SPConv for scales 4 and 2.  
    Although the SPConvs at different scales employ different numbers of filters, the filtered sub-pixels for all scales exhibit significant 2D spatial correlations~(illustrated with color gradients) due to the subsequent $\mathcal{DS}$. 
    Visualized convolution filters trained to capture inter-scale correlations are shown in Figure~\ref{fig:implicitgrids}.
    \vspace{-0.5cm}
  }
  \label{fig:pixel_shuffle}
\end{figure}

For the next step, we investigate the mechanism of the SPConv, a commonly used and efficient scale-specific upsampler. 
Upon detailed visualization of SPConv, we observe that, although SPConv at different scales uses varying numbers of convolution filters, it shares a common goal.
Specifically, it divides each LR pixel~(denoted as a \textit{grid} and illustrated by the black bolded lines in Figure~\ref{fig:pixel_shuffle}) into $\mathit{r}^{2}$ sub-pixels.
Furthermore, these $\mathit{r}^{2}$ sub-pixels exhibit strong inter-scale correlation in 2D space due to the subsequent depth-to-space~($\mathcal{DS}$) operation.
This observation suggests that SPConvs at different scales fundamentally operate in the same way. 
Consequently, SPConvs across all scales can be unified into a single module based on this similarity.

\vspace{-0.1cm}
\subsection{Implicit Grid Convolution}
\vspace{-0.1cm}
We propose IGConv, which integrates SPConv across all scales by parameterizing convolution filters that predict sub-pixels with inter-scale correlations employing hyper-network.
Specifically, the inter-scale correlations denote the size and relative position of sub-pixels that vary with $r$.
IGConv consists of three main components: the hyper-network, convolution operation, and upsampling operation. 
These can be represented as follows:
\begin{equation}
\begin{gathered}
    K = \mathcal{H}(r), \\
    M' = M \ast K, \\
    I^{SR} = \mathcal{DS}(M', r),
\end{gathered}
\end{equation}
\noindent where $\mathcal{H}$ represents the hyper-network that generates the convolution filter $K\in \mathbb{R}^{k \times k \times C_e \times 3 \cdot r^2}$ according to $r$, and $M'\in \mathbb{R}^{H \times W \times 3\cdot r^{2}}$ refers to the feature map obtained by convolving $M$ with $K$. 
$M'$ is then passed through the upsampling operation $\mathcal{DS}(\cdot): \mathbb{R}^{H \times W \times 3 \cdot r^2} \mapsto \mathbb{R}^{rH \times rW \times 3}$, resulting in $I_{SR}$.

Since $\mathcal{H}$ only depends on $r$, it can be pre-computed to targeted scales and excluded during the inference phase, making the instantiated IGConv functionally identical to SPConv.
Furthermore, because IGConv does not add any modules to the utilized encoder, the model with instantiated IGConv maintains the same inference cost and parameters as the scale-specific model employing SPConv.
Even when training, additional computations, and parameters brought by IGConv are negligible compared to those brought by scale-specific encoders and upsamplers.

% \subsubsection{Hyper Network} 
\vspace{0.2cm}
\noindent\textbf{Hyper-Network} 
The hyper-network uses INR-based methods to generate $K$ depending on the inter-scale correlations formulated as:
\vspace{-0.1cm}
\begin{equation}\label{eqn:hypernet}
\begin{gathered}
    F_{r} = h(r), \\
    K = \mathcal{R}(f(F_{r})),
\end{gathered}
\end{equation}
\vspace{-0.1cm}

\noindent where $h$ represents the coefficient estimator that generates Fourier coefficients $F_{r}\in \mathbb{R}^{C_{e}\cdot k^{2}\times r\times r\times C_{h}}$ according to the $r$, and $f(\cdot)\in\mathbb{R}^{C_{e}\cdot k^{2}\times r\times r\times C_{h}} \mapsto \mathbb{R}^{C_{e}\cdot k^{2}\times r\times r\times 3}$ is parameterized MLPs with ReLU activations that predict intermediate representations from $F_{r}$.
Lastly, $\mathcal{R}(\cdot)\in\mathbb{R}^{C_{e}\cdot k^{2}\times r\times r\times 3} \mapsto \mathbb{R}^{k\times k\times C_{e} \times 3r^{2}}$ denotes the reshape operation that converts the predicted intermediate representations into $K$, convolution filters for scale-specific modulation.
Specifically, $F_{r}$ is inferred by the following process:
\begin{equation}\label{eqn:fourier_estimator}
\begin{gathered}
    C_{r} = \langle\delta^{x}_{r}, Z^{x}\rangle + \langle\delta^{y}_{r}, Z^{y}\rangle, s_{r} = h^{s}(2/r), \\
    F_{r} = Z^{amp} \odot \begin{bmatrix}
        \mathrm{cos}(\pi(C_{r} + s_{r})) \\
        \mathrm{sin}(\pi(C_{r} + s_{r}))
    \end{bmatrix},  
\end{gathered}
\end{equation}
\newcommand{\oc}{\cellcolor{orange}}

\begin{table*}[!ht]
\caption{
    Comparisons of fixed-scale upsamplers~(SPConv, SPConv$^{+}$) and our proposed multi-scale upsamplers~(IGConv, IGConv$^{+}$) on various encoders trained on the DIV2K dataset. Results from SPConv{\scriptsize($\times r$)} and SPConv$^{+}${\scriptsize($\times r$)} are measured by each scale-specific model, while results from IGConv and IGConv$^{+}$ are measured by \textbf{a single model}.  The only best result is bolded.
}\label{tab:fixedscale_div2k}
\vspace{-0.2cm}
\setlength{\extrarowheight}{0.2pt}
\resizebox{\textwidth}{!}{%
\begin{tabular}{c|c|ll|ccccc}
Dataset & Scale & \begin{tabular}[c]{@{}l@{}}Encoder \\ [-0.8ex] (Operator)\end{tabular} & Upsampler & Set5 & Set14 & B100 & Urban100 & Manga109 \\ [-0.4ex] \hline
\multirow{27}{*}[5ex]{DIV2K} & \multirow{9}{*}[1.5ex]{2} & \multirow{3}{*}[0.3ex]{\begin{tabular}[c]{@{}l@{}}EDSR~\cite{EDSR}\\ (CNN)\end{tabular}} & SPConv$^{+}${\scriptsize($\times$2)} & 38.11/0.9602 & 33.92/0.9195 & 32.32/0.9013 & 32.93/0.9351 & 39.10/0.9773 \\ [-0.4ex]
 &  &  & IGConv & 38.21/0.9612 & 33.96/0.9209 & 32.34/0.9016 & 32.94/0.9359 & 39.13/0.9780 \\ [-0.4ex]
 &  &  & \textbf{IGConv$^{+}$} & \textbf{38.24}/\textbf{0.9614} & 33.96/0.9209 & 32.34/\textbf{0.9018} & \textbf{33.00}/\textbf{0.9360} & \textbf{39.25}/\textbf{0.9783} \\ [-0.4ex] \cline{3-9} 
 &  & \multirow{3}{*}[0.3ex]{\begin{tabular}[c]{@{}l@{}}HiT-SRF~\cite{HiTSR}\\ (Transformer)\end{tabular}} & SPConv{\scriptsize($\times$2)} & 38.26/0.9615 & 34.01/0.9214 & 32.37/0.9023 & 33.13/0.9372 & 39.47/\textbf{0.9787} \\ [-0.4ex]
 &  &  & IGConv & 38.16/0.9604 & \textbf{34.02}/0.9214 & 32.35/0.9020 & 33.21/0.9377 & 39.34/0.9781 \\ [-0.4ex]
 &  &  & \textbf{IGConv$^{+}$} & \textbf{38.30}/0.9615 & 33.97/0.9210 & \textbf{32.38}/0.9023 & \textbf{33.22}/0.9377 & 39.47/0.9786 \\ [-0.4ex] \cline{3-9} 
 &  & \multirow{3}{*}[0.3ex]{\begin{tabular}[c]{@{}l@{}}MambaIR-lt~\cite{MambaIR}\\ (SSM)\end{tabular}} & SPConv{\scriptsize($\times$2)} & 38.16/0.9610 & 34.00/0.9212 & 32.34/0.9017 & 32.92/0.9356 & 39.31/0.9779 \\ [-0.4ex]
 &  &  & IGConv & 38.20/0.9611 & 34.02/0.9214 & 32.34/0.9014 & 33.02/0.9365 & 39.28/0.9782 \\ [-0.4ex]
 &  &  & \textbf{IGConv$^{+}$} & 38.20/\textbf{0.9613} & \textbf{34.11}/\textbf{0.9221} & \textbf{32.36}/\textbf{0.9019} & \textbf{33.18}/\textbf{0.9372} & \textbf{39.44}/\textbf{0.9786} \\ [-0.4ex] \cline{2-9} 
 & \multirow{9}{*}[1.5ex]{3} & \multirow{3}{*}[0.3ex]{\begin{tabular}[c]{@{}l@{}}EDSR~\cite{EDSR}\\ (CNN)\end{tabular}} & SPConv$^{+}${\scriptsize($\times$3)} & 34.65/0.9280 & 30.52/0.8462 & 29.25/0.8093 & 28.80/0.8653 & 34.17/0.9476 \\ [-0.4ex]
 &  &  & IGConv & 34.70/0.9294 & 30.56/0.8469 & 29.28/0.8097 & 28.90/0.8671 & 34.31/0.9491 \\ [-0.4ex]
 &  &  & \textbf{IGConv$^{+}$} & \textbf{34.74}/\textbf{0.9298} & \textbf{30.65}/\textbf{0.8481} & \textbf{29.30}/\textbf{0.8103} & \textbf{28.95}/\textbf{0.8675} & \textbf{34.47}/\textbf{0.9496}\\ [-0.4ex] \cline{3-9} 
 &  & \multirow{3}{*}[0.3ex]{\begin{tabular}[c]{@{}l@{}}HiT-SRF~\cite{HiTSR}\\ (Transformer)\end{tabular}} & SPConv{\scriptsize($\times$3)} & 34.75/0.9300 & 30.61/0.8475 & 29.29/0.8106 & 28.99/0.8687 & 34.53/0.9502 \\ [-0.4ex]
 &  &  & IGConv & 34.69/0.9292 & 30.60/0.8476 & 29.26/0.8098 & 29.02/\textbf{0.8694} & 34.46/0.9499 \\ [-0.4ex]
 &  &  & \textbf{IGConv$^{+}$} & \textbf{34.78}/\textbf{0.9302} & \textbf{30.69}/\textbf{0.8488} & \textbf{29.32}/\textbf{0.8111} & \textbf{29.06}/0.8693 & \textbf{34.67}/\textbf{0.9506} \\ [-0.4ex] \cline{3-9} 
 &  & \multirow{3}{*}[0.3ex]{\begin{tabular}[c]{@{}l@{}}MambaIR-lt~\cite{MambaIR}\\ (SSM)\end{tabular}} & SPConv{\scriptsize($\times$3)} & 34.72/0.9296 & 30.63/0.8475 & 29.29/0.8099 & 29.00/\textbf{0.8689} & 34.39/0.9495 \\ [-0.4ex]
 &  &  & IGConv & 34.70/0.9294 & 30.59/0.8474 & 29.27/0.8094 & 28.91/0.8672 & 34.37/0.9492 \\ [-0.4ex]
 &  &  & \textbf{IGConv$^{+}$} & \textbf{34.74}/\textbf{0.9298} & \textbf{30.68}/\textbf{0.8487} & \textbf{29.30}/\textbf{0.8105} & \textbf{29.04}/0.8687 & \textbf{34.62}/\textbf{0.9502} \\ [-0.4ex] \cline{2-9} 
 & \multirow{9}{*}[1.5ex]{4} & \multirow{3}{*}[0.3ex]{\begin{tabular}[c]{@{}l@{}}EDSR~\cite{EDSR}\\ (CNN)\end{tabular}} & SPConv$^{+}${\scriptsize($\times$4)} & 32.46/0.8968 & 28.80/0.7876 & 27.71/0.7420 & 26.64/0.8033 & 31.02/0.9148 \\ [-0.4ex]
 &  &  & IGConv & 32.57/0.8990 & 28.84/0.7880 & 27.76/0.7426 & 26.75/0.8060 & 31.29/0.9178 \\ [-0.4ex]
 &  &  & \textbf{IGConv$^{+}$} & \textbf{32.59}/\textbf{0.8996} & \textbf{28.91}/\textbf{0.7890} & \textbf{27.79}/\textbf{0.7433} & \textbf{26.82}/\textbf{0.8064} & \textbf{31.43}/\textbf{0.9182} \\ [-0.4ex] \cline{3-9} 
 &  & \multirow{3}{*}[0.3ex]{\begin{tabular}[c]{@{}l@{}}HiT-SRF~\cite{HiTSR}\\ (Transformer)\end{tabular}} & SPConv{\scriptsize($\times$4)} & 32.55/0.8999 & 28.87/0.7880 & 27.75/0.7432 & 26.80/0.8069 & 31.26/0.9171 \\ [-0.4ex]
 &  &  & IGConv & 32.53/0.8988 & 28.90/0.7887 & 27.71/0.7422 & 26.88/\textbf{0.8085} & 31.31/0.9184 \\ [-0.4ex]
 &  &  & \textbf{IGConv$^{+}$} & \textbf{32.60}/\textbf{0.9001} & \textbf{28.95}/\textbf{0.7892} & \textbf{27.80}/\textbf{0.7440} & \textbf{26.91}/0.8083 & \textbf{31.57}/\textbf{0.9198} \\ [-0.4ex] \cline{3-9} 
 &  & \multirow{3}{*}[0.3ex]{\begin{tabular}[c]{@{}l@{}}MambaIR-lt~\cite{MambaIR}\\ (SSM)\end{tabular}} & SPConv{\scriptsize($\times$4)} & 32.51/0.8993 & 28.85/0.7876 & 27.75/0.7423 & 26.75/0.8051 & 31.26/0.9175 \\ [-0.4ex]
 &  &  & IGConv & 32.50/0.8992 & 28.86/0.7879 & 27.75/0.7422 & 26.72/0.8045 & 31.29/0.9175 \\ [-0.4ex]
 &  &  & \textbf{IGConv$^{+}$} & \textbf{32.62}/\textbf{0.8997} & \textbf{28.93}/\textbf{0.7893} & \textbf{27.80}/\textbf{0.7437} & \textbf{26.87}/\textbf{0.8068} & \textbf{31.51}/\textbf{0.9185} \\ [-0.4ex] \hline
\end{tabular}
}
\vspace{-0.5cm}
\end{table*}

\noindent where $Z^{amp}\in \mathbb{R}^{C_{h}}$ denotes the scale-invariant latent code, $C_{r}\in\mathbb{R}^{C_{e}\cdot k^{2}\times r\times r\times C_{h}}$ refers to the coordinate matrix representing the relative coordinates according to the $r$, and $s_{r}$ represents the size according to the $r$.
$F_{r}$ is created by element-wisely multiplying the scale-variant Fourier matrix, formed by $C_{r}$ and $s_{r}$, with $Z^{amp}$.
$C_{r}$ is generated by matrix multiplicating the uniformly sampled 2D regular coordinates $\delta^{x}_{r}, \delta^{y}_{r}\in [-1, 1]^{1 \times r \times r \times 1}$ with the scale-invariant latent codes $Z^{x}, Z^{y}\in \mathbb{R}^{C_{e} \cdot k^{2} \times 1 \times 1 \times C_{h}/2}$ respectively, and then summing the results.
$s_{r}$ is generated by feeding the reciprocal of the $r$, proportional to the size to be predicted sub-pixels, into a single linear layer~($h^{s}$).

In summary, the convolution filters are estimated from the size and regular coordinates evenly distributed by $r\times r$ in 2D space.
These attributes correspond to the size and coordinates of $r^{2}$ filtered sub-pixels in each grid after $\mathcal{DS}$. 
Since filtered sub-pixels of SPConv at any scale can be represented in the same way, $\mathcal{H}$ can effectively predict convolutional filters at any integer scale for upsampling.

\vspace{-0.1cm}
\subsection{Frequency Loss}
\vspace{-0.1cm}
Mapping signals employing MLPs induces spectral bias~\cite{LTE}. 
Therefore, in addition to the commonly used pixel-wise L1 loss, we leverage frequency loss~\cite{MAXIM, ShuffleMixer} to make the model focus on high-frequency detail, as follows:

\vspace{-0.1cm}
\begin{gather}\label{eqn:TotalLoss}
    \mathcal{L} = ||I^{HR} - I^{SR}||_1 + \lambda||\mathcal{F}(I^{HR}) - \mathcal{F}(I^{SR})||_1,
\end{gather}
\vspace{-0.1cm}

\noindent where $\mathcal{F}$ denotes the Fast Fourier transform, and $\lambda$ is a weight parameter set to be 0.05 empirically.

\subsection{Implicit Grid Sampling}
SPConv's performance is limited since it upscales the $M$ without utilizing the rich representation from it. 
For this reason, many SR studies focusing on performance improvements~\cite{EDSR, MambaIR, SRFormer, ATD, HAT} employ SPConv$^{+}$, which leverages extra convolution after $\mathcal{DS}$~(see Figure~\ref{fig:igconvvis} (a)). 
However, the extra convolution in high-resolution~(HR) space brings significant computational overhead, increasing SPConv$^{+}$'s latency nearly 10$\times$ over SPConv, as shown in Table~\ref{tab:efficiency}.
To address this limitation, we propose an IGSample inspired by previous research aimed at upsampling feature maps~\cite{DySample}, as formulated below:
\begin{equation}
\begin{gathered}
    K^{o}, K^{s} = \mathcal{H}_{\mathcal{S}}(r), \\ 
    \delta^{xy} = (M \ast K^{o}) \odot 0.5\sigma(M \ast K^{s}), \\ 
    \mathrm{x}_{r} = \mathrm{x}^{bi}_{r} + \mathcal{DS}(\delta^{xy}, r), \\ 
    I^{\uparrow} = \mathcal{S}(I^{LR}, \mathrm{x}_{r}), 
\end{gathered}  
\end{equation}

\begin{table*}[!ht]
\caption{
    Comparisons of fixed-scale upsamplers~(SPConv, SPConv$^{+}$) and our proposed multi-scale upsamplers~(IGConv, IGConv$^{+}$) on various encoders trained on the DF2K dataset. Results from SPConv{\scriptsize($\times r$)} and SPConv$^{+}${\scriptsize($\times r$)} are measured by each scale-specific model, while results from IGConv and IGConv$^{+}$ are measured by \textbf{a single model}. The only best result is bolded.
}\label{tab:fixedscale_df2k}
% \vspace{-0.2cm}
% \renewcommand{\arraystretch}{0.7}
\setlength{\extrarowheight}{0.2pt}
\resizebox{\textwidth}{!}{%
\begin{tabular}{c|c|ll|ccccc}
Dataset & Scale & \begin{tabular}[c]{@{}l@{}}Encoder \\ [-0.8ex] (Operator)\end{tabular} & Upsampler & Set5 & Set14 & B100 & Urban100 & Manga109 \\ [-0.4ex] \hline
\multirow{27}{*}[5ex]{DF2K} & \multirow{9}{*}[1.5ex]{2} & \multirow{3}{*}[0.3ex]{\begin{tabular}[c]{@{}l@{}}SMFANet+~\cite{SMFANet}\\ (CNN)\end{tabular}} & SPConv
{\scriptsize($\times$2)} & \textbf{38.19}/0.9611 & 33.92/0.9207 & 32.32/\textbf{0.9015} & 32.70/0.9331 & \textbf{39.46}/\textbf{0.9787} \\ [-0.4ex]
 &  &  & IGConv & 38.16/0.9610 & \textbf{33.96}/\textbf{0.9213} & 32.32/0.9014 & 32.73/0.9332 & 39.38/0.9785 \\ [-0.4ex]
 &  &  & \textbf{IGConv$^{+}$} & 38.14/0.9611 & 33.92/0.9208 & 32.32/0.9014 & \textbf{32.76}/\textbf{0.9334} & 39.40/0.9786 \\ [-0.4ex] \cline{3-9} 
 &  & \multirow{3}{*}[0.3ex]{\begin{tabular}[c]{@{}l@{}}SRFormer~\cite{SRFormer}\\ (Transformer)\end{tabular}} & SPConv$^{+}${\scriptsize($\times$2)} & 38.51/\textbf{0.9627} & 34.44/0.9253 & 32.57/0.9046 & 34.09/0.9449 & \textbf{40.07}/\textbf{0.9802} \\ [-0.4ex]
 &  &  & IGConv & 38.44/0.9625 & 34.64/0.9267 & 32.56/0.9048 & 34.28/0.9462 & 39.88/0.9798 \\ [-0.4ex]
 &  &  & \textbf{IGConv$^{+}$} & \textbf{38.53}/0.9626 & \textbf{34.72}/\textbf{0.9268} & \textbf{32.60}/\textbf{0.9052} & \textbf{34.42}/\textbf{0.9468} & 40.03/0.9797 \\ [-0.4ex] \cline{3-9} 
 &  & \multirow{3}{*}[0.3ex]{\begin{tabular}[c]{@{}l@{}}MambaIR~\cite{MambaIR}\\ (SSM)\end{tabular}} & SPConv$^{+}${\scriptsize($\times$2)} & \textbf{38.57}/\textbf{0.9627} & 34.67/0.9261 & 32.58/0.9048 & 34.15/0.9446 & \textbf{40.28}/\textbf{0.9806} \\ [-0.4ex]
 &  &  & IGConv & 38.48/0.9624 & 34.68/0.9264 & 32.58/0.9047 & 34.26/0.9453 & 40.14/0.9803 \\ [-0.4ex]
 &  &  & \textbf{IGConv$^{+}$} & 38.55/0.9625 & \textbf{34.81}/\textbf{0.9270} & \textbf{32.62}/\textbf{0.9052} & \textbf{34.37}/\textbf{0.9461} & 40.19/0.9802 \\ [-0.4ex] \cline{2-9} 
 & \multirow{9}{*}[1.5ex]{3} & \multirow{3}{*}[0.3ex]{\begin{tabular}[c]{@{}l@{}}SMFANet+~\cite{SMFANet}\\ (CNN)\end{tabular}} & SPConv{\scriptsize($\times$3)} & \textbf{34.66}/\textbf{0.9292} & 30.57/0.8461 & 29.25/0.8090 & \textbf{28.67}/\textbf{0.8611} & 34.45/0.9490 \\ [-0.4ex]
 &  &  & IGConv & 34.62/0.9290 & 30.56/0.8461 & 29.25/0.8090 & 28.64/0.8606 & 34.45/0.9490 \\ [-0.4ex]
 &  &  & \textbf{IGConv$^{+}$} & 34.58/0.9287 & \textbf{30.58}/\textbf{0.8464} & 29.24/0.8089 & 28.66/0.8610 & 34.45/0.9490 \\ [-0.4ex] \cline{3-9} 
 &  & \multirow{3}{*}[0.3ex]{\begin{tabular}[c]{@{}l@{}}SRFormer~\cite{SRFormer}\\ (Transformer)\end{tabular}} & SPConv$^{+}${\scriptsize($\times$3)} & 35.02/0.9323 & 30.94/0.8540 & 29.48/0.8156 & 30.04/0.8865 & 35.26/0.9543 \\ [-0.4ex]
 &  &  & IGConv & 34.96/0.9323 & 30.95/0.8543 & 29.47/0.8157 & 30.11/0.8876 & 35.16/0.9543 \\ [-0.4ex]
 &  &  & \textbf{IGConv$^{+}$} & \textbf{35.08}/\textbf{0.9329} & \textbf{31.06}/\textbf{0.8551} & \textbf{29.52}/\textbf{0.8166} & \textbf{30.25}/\textbf{0.8888} & \textbf{35.45}/\textbf{0.9550} \\ [-0.4ex] \cline{3-9} 
 &  & \multirow{3}{*}[0.3ex]{\begin{tabular}[c]{@{}l@{}}MambaIR~\cite{MambaIR}\\ (SSM)\end{tabular}} & SPConv$^{+}${\scriptsize($\times$3)} & 35.08/0.9323 & 30.99/0.8536 & 29.51/0.8157 & 29.93/0.8841 & 35.43/0.9546 \\ [-0.4ex]
 &  &  & IGConv & 35.04/0.9320 & 31.01/0.8535 & 29.50/0.8154 & 29.95/0.8844 & 35.44/0.9545 \\ [-0.4ex]
 &  &  & \textbf{IGConv$^{+}$} & \textbf{35.10}/\textbf{0.9325} & \textbf{31.14}/\textbf{0.8550} & \textbf{29.55}/\textbf{0.8164} & \textbf{30.11}/\textbf{0.8864} & \textbf{35.55}/\textbf{0.9549} \\ [-0.4ex] \cline{2-9} 
 & \multirow{9}{*}[1.5ex]{4} & \multirow{3}{*}[0.3ex]{\begin{tabular}[c]{@{}l@{}}SMFANet+~\cite{SMFANet}\\ (CNN)\end{tabular}} & SPConv{\scriptsize($\times$4)} & 32.51/0.8985 & \textbf{28.87}/\textbf{0.7872} & 27.74/0.7412 & \textbf{26.56}/\textbf{0.7976} & 31.29/\textbf{0.9163} \\ [-0.4ex]
 &  &  & IGConv & 32.47/0.8982 & 28.84/0.7866 & 27.74/0.7413 & 26.54/0.7969 & 31.28/0.9158 \\ [-0.4ex]
 &  &  & \textbf{IGConv$^{+}$} & \textbf{32.52}/\textbf{0.8988} & 28.83/0.7867 & 27.74/0.7413 & 26.55/0.7974 & 31.29/0.9161 \\ [-0.4ex] \cline{3-9} 
 &  & \multirow{3}{*}[0.3ex]{\begin{tabular}[c]{@{}l@{}}SRFormer~\cite{SRFormer}\\ (Transformer)\end{tabular}} & SPConv$^{+}${\scriptsize($\times$4)} & 32.93/0.9041 & 29.08/0.7953 & 27.94/0.7502 & 27.68/0.8311 & 32.21/0.9271 \\ [-0.4ex]
 &  &  & IGConv & 32.87/0.9046 & 29.08/0.7952 & 27.91/0.7499 & 27.79/0.8333 & 32.14/0.9274 \\ [-0.4ex]
 &  &  & \textbf{IGConv$^{+}$} & \textbf{33.04}/\textbf{0.9047} & \textbf{29.22}/\textbf{0.7971} & \textbf{27.99}/\textbf{0.7509} & \textbf{27.93}/\textbf{0.8350} & \textbf{32.45}/\textbf{0.9288} \\ [-0.4ex] \cline{3-9} 
 &  & \multirow{3}{*}[0.3ex]{\begin{tabular}[c]{@{}l@{}}MambaIR~\cite{MambaIR}\\ (SSM)\end{tabular}} & SPConv$^{+}${\scriptsize($\times$4)} & 33.03/\textbf{0.9046} & 29.20/0.7961 & 27.98/0.7503 & 27.68/0.8287 & 32.32/0.9272 \\ [-0.4ex]
 &  &  & IGConv & 32.98/0.9041 & 29.17/0.7955 & 27.97/0.7498 & 27.68/0.8288 & 32.36/0.9271 \\ [-0.4ex]
 &  &  & \textbf{IGConv$^{+}$} & \textbf{33.05}/0.9045 & \textbf{29.25}/\textbf{0.7969} & \textbf{28.02}/\textbf{0.7512} & \textbf{27.80}/\textbf{0.8314} & \textbf{32.52}/\textbf{0.9280} \\ [-0.4ex] \hline
\end{tabular}
}
% \vspace{-0.3cm}
\end{table*}

\noindent where $\mathcal{H}_{\mathcal{S}}$ denotes hyper-network to predict convolution filters $K^{o}$, $K^{s}\in\mathbb{R}^{k\times k\times C_{e} \times 6\cdot r^{2}}$ depending on $\mathit{r}$ similar with the $\mathcal{H}$ in IGConv.
$K^{o}$ and $K^{s}$ are convolution filters to predict the 2D direction and constraint scope for bilinear upsampling~($\mathcal{S}$) on each RGB space.
After applying a sigmoid~($\sigma$) and multiplying 0.5 to the predicted scope, it is multiplied element-wise with the direction to generate calibrating offset $\delta^{xy}\in\mathbb{R}^{H\times W\times 6\cdot r^{2}}$.
Subsequently, $\delta^{xy}$ is upsampled through $\mathcal{DS}$, then added to $\mathrm{x}^{bi}_{r} \in \mathbb{R}^{rH \times rW \times 6}$, which indicates the coordinates for $\mathcal{S}$ that is repeated three times channel-wise to represent each RGB space.
Finally, the upsampled image $I^{\uparrow}$ is created by performing $\mathcal{S}$ from $I_{LR}$ in each RGB space based on $\mathrm{x}_{r}$ and then added to $I^{SR}$. 

As a result, our IGSample upsamples $I^{LR}$ input-dependently by adjusting the coordinates for $\mathcal{S}$ leveraging the rich information from $M$.
IGSample also reduces spectral bias by adding low-frequency biased upsampled image~\cite{LTE} to $I^{SR}$.

\subsection{Feature-level Geometric Re-param.}
We propose FGRep inspired by input-level geometric ensemble~\cite{EDSR} and feature-level local ensemble~\cite{LIIF} to improve performance by enabling ensemble prediction, defined as follows:
\begin{equation}\label{eqn:geo_case1}
    M' = \frac{1}{8} \sum_{i=1}^{8} \mathcal{A}^{-1}_{i}(\mathcal{A}_{i}(M) \ast K),
\end{equation}
\noindent where $\mathcal{A}$ refers to augmentation functions that consist of 8 transformations, including flip, rotation, and identity.
% FGRep performs each $\mathcal{A}$ on $M$ to create augmented versions of $M$, followed by convolution with $K$.
Each $\mathcal{A}$ applies on $M$ to create an augmented version of $M$, followed by convolution with $K$.
Then, inverse augmentation $\mathcal{A}^{-1}$ is applied to each filtered output to revert them to their original state, and all filtered outputs are averaged to produce $M'$.
The $\mathcal{H}$, and $\mathcal{DS}$ in IGConv are omitted.

This is similar to the local ensemble as it performs the ensemble on the final feature $M$ and is also similar to the geometric ensemble in how augmentations are applied.
Interestingly, performing convolution on augmented feature maps with a single kernel followed by inverse augmentation is equivalent to applying convolution to a single feature map with augmented kernels, leading to the redefinition of Equation~\ref{eqn:geo_case1} as follows:
\begin{equation}\label{eqn:geo_case2}
    M' = \frac{1}{8} \sum_{i=1}^{8} M * \mathcal{A}_{i}(K).
\end{equation}
Furthermore, performing convolution on a single feature map with multiple kernels and then summing up results can be converted to performing convolution on a single feature map with a single kernel via structural re-parameterization~\cite{ACNet}. 
Equation~\ref{eqn:geo_case2} is redefined by structural re-parameterization as:
\begin{equation}
\begin{gathered}\label{eqn:geo_case3}
    M' = M \ast \Bar{K}, \\
    \mathbf{where} \; \Bar{K} = \frac{1}{8}\sum_{i=1}^{8} \mathcal{A}_{i}(K). 
\end{gathered}
\end{equation}
Consequently, FGRep allows the upsampler to produce ensembled predictions with only a single forward pass during the inference phase.
We apply FGRep to every kernel predicted by the hyper-networks ($K, K^{o}, K^{s}$).

\section{Experiments}
% This section describes the experimental setting and results of our methods. 
% % 우리는 Pytorch와 BasicSR toolbox를 바탕으로 구현 및 실험한다.
% We 
% \ro{We will release our code to the public on GitHub if the paper is accepted.}

\begin{table*}[!ht]
\caption{
    Comparisons of SPConv$^{+}$ and IGConv$^{+}$ on the methods adopting the pre-training strategy. 
    Results from SPConv$^{+}${\scriptsize($\times r$)} are measured by each scale-specific model, while results from IGConv$^{+}$ are measured by \textbf{a single model}.
    The best and the second-best results are bolded and underlined, respectively.
    % \fcolorbox{orange}{orange}{\textcolor{orange}{\rule{1ex}{1ex}}} denotes methods employ multi-scale framework.
}\label{tab:large_scale_models}
% \vspace{-0.2cm}
% \renewcommand{\arraystretch}{0.9}
% \setlength{\extrarowheight}{0.2pt}
\resizebox{\textwidth}{!}{%
\begin{tabular}{c|c|ll|ccccc}
Pre-train & Scale & Encoder & Upsampler & Set5 & Set14 & B100 & Urban100 & Manga109 \\ \Xhline{2\arrayrulewidth}
\multirow{2}{*}{DF2K} & \multirow{4}{*}{$\times$2} & \multirow{2}{*}{ATD~\cite{ATD}} & SPConv$^{+}${\scriptsize($\times$2)} & 38.61/0.9629 & 34.92/0.9275 & 32.64/0.9054 & 34.73/0.9476 & 40.35/\underline{0.9810} \\
 &  & & \textbf{IGConv$^{+}$} & \underline{38.68}/\underline{0.9631} & 35.00/\underline{0.9280} & \underline{32.69}/\underline{0.9059} & \underline{34.94}/\underline{0.9491} & 40.29/0.9804 \\ \cline{1-1} \cline{3-9} 
% \multirow{4}{*}{ImageNet} &  & IPT~\cite{IPT} & SPConv$^{+}${\scriptsize($\times$2)} & 38.37/- & 34.43/- & 32.48/- & 33.76/- & -/- \\
%  & & EDT~\cite{EDT} & SPConv$^{+}${\scriptsize($\times$2)} & 38.63/0.9632 & 34.80/0.9273 & 32.62/0.9052 & 34.27/0.9456 & 40.37/\underline{0.9811} \\
\multirow{2}{*}{ImageNet} &  & \multirow{2}{*}{HAT~\cite{HAT}} & SPConv$^{+}${\scriptsize($\times$2)} & \textbf{38.73}/\textbf{0.9637} & \underline{35.13}/\textbf{0.9282} & \underline{32.69}/\textbf{0.9060} & 34.81/0.9489 & \textbf{40.71}/\textbf{0.9819} \\
 % & & HAT-L~\cite{HAT} & SPConv$^{+}$ & 38.91/0.9646 & 35.29/0.9293 & 32.74/0.9066 & 35.09/0.9505 & 41.01/0.9831 \\
 & & & \textbf{IGConv$^{+}$} & \underline{38.68}/\underline{0.9631} & \textbf{35.16}/\textbf{0.9282} & \textbf{32.71}/\textbf{0.9060} & \textbf{34.98}/\textbf{0.9494} & \underline{40.39}/0.9809 \\ \Xhline{2\arrayrulewidth}
 % & & HAT-L & \textbf{IGConv$^{+}$} &  &  &  &  &  \\ \Xhline{2\arrayrulewidth}
\multirow{2}{*}{DF2K} & \multirow{4}{*}{$\times$3} & \multirow{2}{*}{ATD~\cite{ATD}} & SPConv$^{+}${\scriptsize($\times$3)} & 35.15/0.9331 & 31.15/0.8556 & 29.58/0.8175 & 30.52/0.8924 & 35.64/0.9558 \\
 & & & \textbf{IGConv$^{+}$} & \textbf{35.17}/\underline{0.9334} & 31.22/\underline{0.8564} & \underline{29.61}/\textbf{0.8183} & \underline{30.76}/0.8946 & \underline{35.84}/0.9565 \\ \cline{1-1} \cline{3-9}  
% \multirow{4}{*}{ImageNet} &  & IPT~\cite{IPT} & SPConv$^{+}${\scriptsize($\times$3)} & 34.87/- & 30.85/- & 29.38/- & 29.49/- & -/- \\
%  &  & EDT~\cite{EDT} & SPConv$^{+}${\scriptsize($\times$3)} & 35.13/0.9328 & 31.09/0.8553 & 29.53/0.8165 & 30.07/0.8863 & 35.47/0.9550 \\
\multirow{2}{*}{ImageNet} &  & \multirow{2}{*}{HAT~\cite{HAT}} & SPConv$^{+}${\scriptsize($\times$3)} & \underline{35.16}/\textbf{0.9335} & \underline{31.33}/\textbf{0.8576} & 29.59/0.8177 & 30.70/\underline{0.8949} & \underline{35.84}/\underline{0.9567} \\
 % &  & HAT-L~\cite{HAT} & SPConv$^{+}$ & 35.28/0.9345 & 31.47/0.8584 & 29.63/0.8191 & 30.92/0.8981 & 36.02 0.9576 \\
 &  &  & \textbf{IGConv$^{+}$} & 35.13/\textbf{0.9335} & \textbf{31.46}/\textbf{0.8576} & \textbf{29.62}/\underline{0.8182} & \textbf{30.78}/\textbf{0.8951} & \textbf{35.95}/\textbf{0.9568} \\ \Xhline{2\arrayrulewidth}
 % &  & HAT-L & \textbf{IGConv$^{+}$} &  &  &  &  &  \\ \Xhline{2\arrayrulewidth}
\multirow{2}{*}{DF2K} & \multirow{4}{*}{$\times$4} & \multirow{2}{*}{ATD~\cite{ATD}} & SPConv$^{+}${\scriptsize($\times$4)} & 33.14/0.9061 & 29.25/0.7976 & 28.02/0.7524 & 28.22/0.8414 & 32.65/0.9308 \\
 &  & & \textbf{IGConv$^{+}$} & 33.13/0.9061 & 29.36/0.7994 & \underline{28.07}/\textbf{0.7536} & \underline{28.43}/0.8444 & \underline{32.92}/\underline{0.9319} \\ \cline{1-1} \cline{3-9}  
% \multirow{4}{*}{ImageNet} &  & IPT~\cite{IPT} & SPConv$^{+}${\scriptsize($\times$4)} & 32.64/- & 29.01/- & 27.82/- & 27.26/- & -/- \\
%  &  & EDT~\cite{EDT} & SPConv$^{+}${\scriptsize($\times$4)} & 32.82/0.9031 & 29.09/0.7939 & 27.91/0.7483 & 27.46/0.8246 & 32.05/0.9254 \\
\multirow{2}{*}{ImageNet} &  & \multirow{2}{*}{HAT~\cite{HAT}} & SPConv$^{+}${\scriptsize($\times$4)} & \textbf{33.18}/\underline{0.9073} & \underline{29.38}/\underline{0.8001} & 28.05/\underline{0.7534} & 28.37/\underline{0.8447} & 32.87/\underline{0.9319} \\
 % &  & HAT-L~\cite{HAT} & SPConv$^{+}$ & 33.30 0.9083 & 29.47/0.8015 & 28.09/0.7551 & 28.60/0.8498 & 33.09 0.9335 \\
 &  & & \textbf{IGConv$^{+}$} & \underline{33.17}/\textbf{0.9074} & \textbf{29.48}/\textbf{0.8008} & \textbf{28.08}/0.7533 & \textbf{28.45}/\textbf{0.8450} & \textbf{33.09}/\textbf{0.9327} \\ \Xhline{2\arrayrulewidth}
 % &  & HAT-L & \textbf{IGConv$^{+}$} &  &  &  &  &  \\ \Xhline{2\arrayrulewidth}
\end{tabular}%
}
% \vspace{-0.2cm}
\end{table*}

\subsection{Training Strategy}
This section describes the training strategy to train multiple integer scales simultaneously.
We randomly sample a scale from $r\in\{2, 3, 4\}$ for each batch, commonly used scales in SR tasks.
After that, we crop a patch ($I_{HR}$) from a high-quality image ($I_{GT}$) to the size of the training patch multiplied by the sampled scale.
$I_{HR}$ is then bicubic downsampled by the randomly sampled scale to create $I_{LR}$, and the model is trained to reconstruct $I_{HR}$ from $I_{LR}$.
Since IGConv can only upsample a single scale per batch, we optimize the model by utilizing generalized gradients averaged across multiple sub-batches.
This approach can be implemented by gradient accumulation or distributed learning, and we use distributed learning with 4 GPUs.
\textit{In all cases, we train multi-scale simultaneously employing IGConv or IGConv$^{+}$ with only the training budget that existing methods employing SPConv or SPConv$^{+}$ used for a single scale, thereby reducing training budget~(training time or GPU demands) by one-third.
}

\subsection{Implemtation Details}
In this section, we describe the implementation details of our proposal methods. 
The $f$ of $\mathcal{H}$ and $\mathcal{H}_{\mathcal{S}}$ are composed of 256 and 128 dimensions, respectively, with four and two hidden layers.
Additionally, $C_{h}$ for $\mathcal{H}$ and $\mathcal{H}_{\mathcal{S}}$ are also set to 256 and 128, respectively.
In practice, since all intermediate representations in $\mathcal{H}$ and $\mathcal{H_{S}}$ are suitably structured for convolution, we implement the $f$ employing 1$\times$1 convolutions.
In all cases, $k$ is set to 3.
We implement our codes based on Pytorch and BasicSR toolbox.
% ~\cite{BasicSR}, and \ro{and is available on Github: https://github.com/dslisleedh/IGConv}.

\subsection{Quantitative Results} 
% \subsubsection{Comparisons on fixed-scale methods}
To validate the importance of multi-scale training and the superiority of our proposed methods, we compare IGConv and IGConv$^{+}$ with SPConv and SPConv$^{+}$ on various encoders~(EDSR~\cite{EDSR}, SMFANet~\cite{SMFANet}, HiT-SRF~\cite{HiTSR}, SRFormer~\cite{SRFormer}, MambaIR~\cite{MambaIR}) with various core operators~(convolution, self-attention, and state-space model), datasets~(DIV2K~\cite{DIV2K} and DF2K~\cite{DF2KDataset}), and upsampler's complexity~(SPConv, and SPConv$^{+}$), respectively. 
For evaluation, we use five commonly used datasets~(Set5 ~\cite{Set5}, Set14 ~\cite{Set14}, B100 ~\cite{BSD100}, Urban100 ~\cite{Urban100}, and Manga109 ~\cite{Manga109}), and measure Peak Signal to Noise Ratio~(PSNR) and Structural Similarity Index Measure~(SSIM) in the y-channel after cropping image's boundary equivalent to the each $r$. 
The training details are provided in the Appendix~\ref{sec:training_details}.

\begin{table}[th]
\caption{
    Comparisons on SPConv, SPConv$^{+}$, IGConv, and IGConv$^{+}$ on efficiency measures after instantiating our methods for the targeted scales. Metrics are calculated by reconstructing an HD~(1280$\times$720) image employing the encoder with 64 channels at an A6000 GPU.
}\label{tab:efficiency}
% \vspace{-0.3cm}
\resizebox{\columnwidth}{!}{%
\begin{tabular}{l|ccc|ccc|ccc}
\multirow{2}{*}{Upsampler} & \multicolumn{3}{c|}{Latency~(ms)} & \multicolumn{3}{c|}{Parameters~(K)} & \multicolumn{3}{c}{Memory~(mb)} \\ 
 & $\times$2 & $\times$3 & $\times$4 & $\times$2 & $\times$3 & $\times$4 & $\times$2 & $\times$3 & $\times$4 \\ \hline
SPConv & \multirow{2}{*}{0.41} & \multirow{2}{*}{0.23} & \multirow{2}{*}{0.25} & \multirow{2}{*}{6.9} & \multirow{2}{*}{15.6} & \multirow{2}{*}{27.7} & \multirow{2}{*}{77.4} & \multirow{2}{*}{46.1} & \multirow{2}{*}{38.9} \\
IGConv$_{inst}$ &  &  &  &  &  &  &  &  &  \\ \hline
SPConv$^{+}$ & 4.14 & 3.92 & 4.90 & 149.4 & 334.1 & 297.2 & 508.8 & 475.5 & 467.2 \\
IGConv$^{+}_{inst}$ & 1.97 & 1.50 & 1.47 & 34.6 & 77.8 & 138.2 & 197.9 & 166.1 & 156.0 \\ \hline
\end{tabular}
}
% \vspace{-0.4cm}
\end{table}

In Table~\ref{tab:fixedscale_div2k}, and Table~\ref{tab:fixedscale_df2k}, IGConv maintains comparable performance to SPConv and SPConv$^{+}$ while reducing training budget and stored parameters by one-third, indicating that scale-specific training is not essential.
Moreover, SRFormer-IGConv$^{+}$ outperforms SRFormer-SPConv$^{+}$ by 0.25dB at Urban100$\times$4, highlighting the superior performance of IGConv$^{+}$.
This result demonstrates that the additional methods~(frequency loss, IGSample, and FGRep) introduced in IGConv$^{+}$ contribute significantly to this performance improvement. 
Ablation studies for the proposed methods can be found in Appendix~\ref{sec:ablation}.

We further validate IGConv$^{+}$ on the methods larger models adopting pre-training strategies~\cite{ATD, HAT}.
As detailed in Table~\ref{tab:large_scale_models}, IGConv$^{+}$ outperforms SPConv$^{+}$ on Manga109$\times$4 employing ATD and HAT as encoders, improving PSNR by 0.27 dB and 0.22dB, respectively.
These results suggest that our multi-scale framework improves performance even on larger models or complex training settings including pre-training and fine-tuning, which aligns with recent research trends. 

We also compare our methods with SPConv and SPConv$^{+}$ on efficiency metrics.
As shown in Table~\ref{tab:efficiency}, IGConv$_{inst}$ exhibits the same computational cost as SPConv while reducing substantial training budget and stored parameters.
Notably, IGConv$^{+}_{inst}$ demonstrates less latency, parameters, and memory usage than SPConv$^{+}$ since all computations are computed in LR space, highlighting our method's remarkable efficiency. 

Finally, we compare SPConv$^{+}$ and IGConv$^{+}$ on training efficiency measures.
As shown in Table~\ref{tab:training_efficiency}, IGConv$^{+}$ significantly reduces training time by three-fold since it only leverages the training budget for SPConv$^{+}$ to train a single scale.
Furthermore, IGConv$^{+}$ also reduces the number of parameters by one-third confirming that additional parameters brought by IGConv$^{+}$ are negligible to those brought by scale-specific encoders and upsamplers. 
Note that additional parameters brought by IGConv$^{+}$ can be further reduced by instantiating IGConv$^{+}$ at the inference phase.
These results demonstrate that our multi-scale framework significantly reduces training overheads, highlighting our proposal's exceptional training efficiency.

\vspace{-0.1cm}
\begin{table}[ht!]
\caption{
    Comparisons of SPConv$^{+}$ and IGConv$^{+}$ on training efficiency metrics.
    All training efficiency metrics are obtained by training HAT~\cite{HAT} using four A6000 GPUs.
    PT and FT denote pre-training on the ImageNet dataset and fine-tuning on the DF2K dataset. \#Params indicates the number of parameters \textit{measured during training phase} including both encoder and upsampler.
    }\label{tab:training_efficiency}
% \vspace{-0.3cm}
\resizebox{\columnwidth}{!}{%
\begin{tabular}{ll|cccl|c}
\multicolumn{2}{l|}{\multirow{2}{*}{Metrics}} & \multicolumn{4}{c|}{SPConv$^{+}$} & \multirow{2}{*}{IGConv$^{+}$} \\
\multicolumn{2}{l|}{} & $\times$2 & $\times$3 & $\times$4 & =~Total &  \\ \hline
\multicolumn{1}{l|}{\multirow{2}{*}{Time~(h)}} & PT & 210 & 210 & 210 & =~630 & \textbf{213} \\
\multicolumn{1}{l|}{} & FT & 68 & 68 & 68 & =~204 & \textbf{69} \\ \hline
\multicolumn{2}{l|}{\#Params~(M)} & 21 & 21 & 21 & =~63 & \textbf{22} \\ \hline
\end{tabular}%
}
% \vspace{-0.4cm}
\end{table}

% \begin{table}[]
% \resizebox{\columnwidth}{!}{%
% \begin{tabular}{l|cccc|c}
% \multirow{2}{*}{} & \multicolumn{4}{c|}{SPConv$^{+}$} & \multirow{2}{*}{IGConv$^{+}$} \\
%  & X2 & X3 & X4 & Total &  \\ \hline
% Time & 241 & 241 & 241 &  & 241 \\
% Params &  &  &  &  &  \\ \hline
% \end{tabular}%
% }
% \end{table}

\subsection{Analysis on Inter-Scale Correlations}
\begin{figure}[h]
  \centering
  \includegraphics[width=\columnwidth]{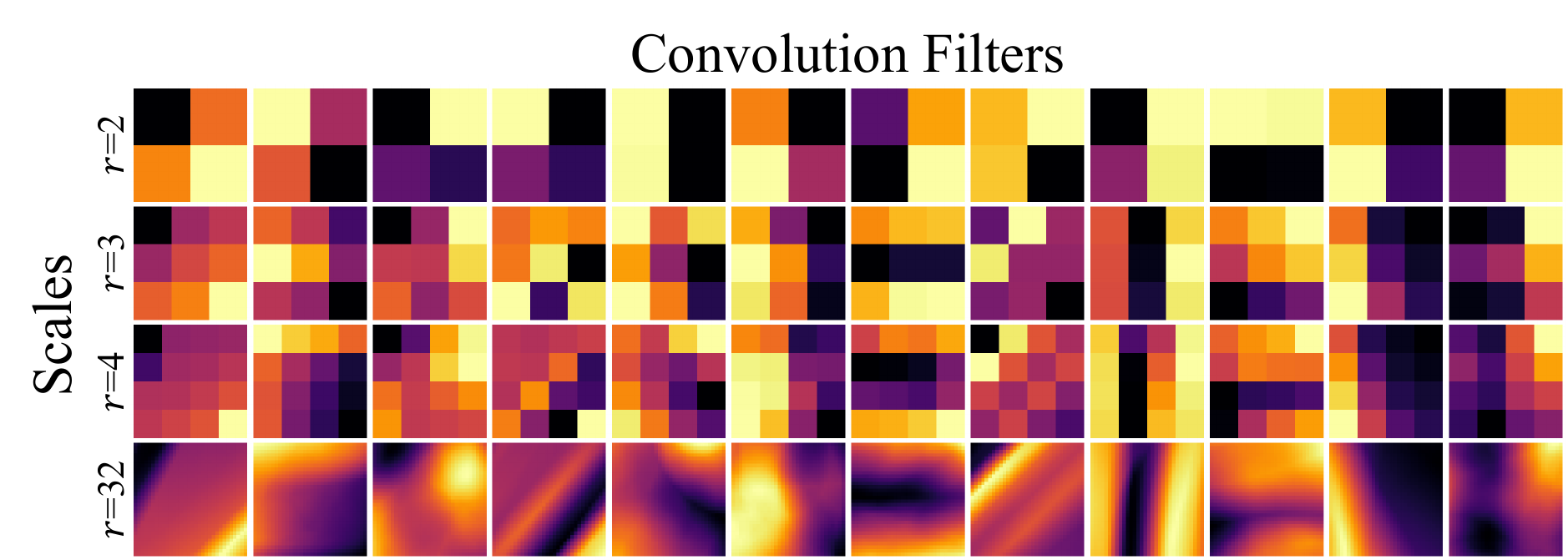}
  \caption{
    Visualizations of 12 convolution filters in front inferred by $\mathcal{H}$ of RDN-IGConv$^{+}$ for scales $\times$2, $\times$3, $\times$4, and $\times$32. 
    More visualizations are provided in the Appendix~\ref{sec:igvis_full}.
  }
  % \vspace{-0.5cm}
  \label{fig:implicitgrids}
\end{figure}

To illustrate the impact of $\mathcal{H}$ that maps inter-scale correlations~(the size and coordinates of predicted sub-pixels) to convolutional filters, we visualise the convolutional filter in RDN-IGConv$^{+}$ at various scales~($\times$2, $\times$3, $\times$4, and $\times$32). 
As shown in Figure~\ref{fig:implicitgrids}, the convolution filters at all scales change continuously in response to the change of $r$, indicating that $\mathcal{H}$ effectively mapped the inter-scale correlation to the convolution filters.

% \vspace{-0.2cm}

\begin{figure}[h!]
  \centering
  \includegraphics[width=\columnwidth]{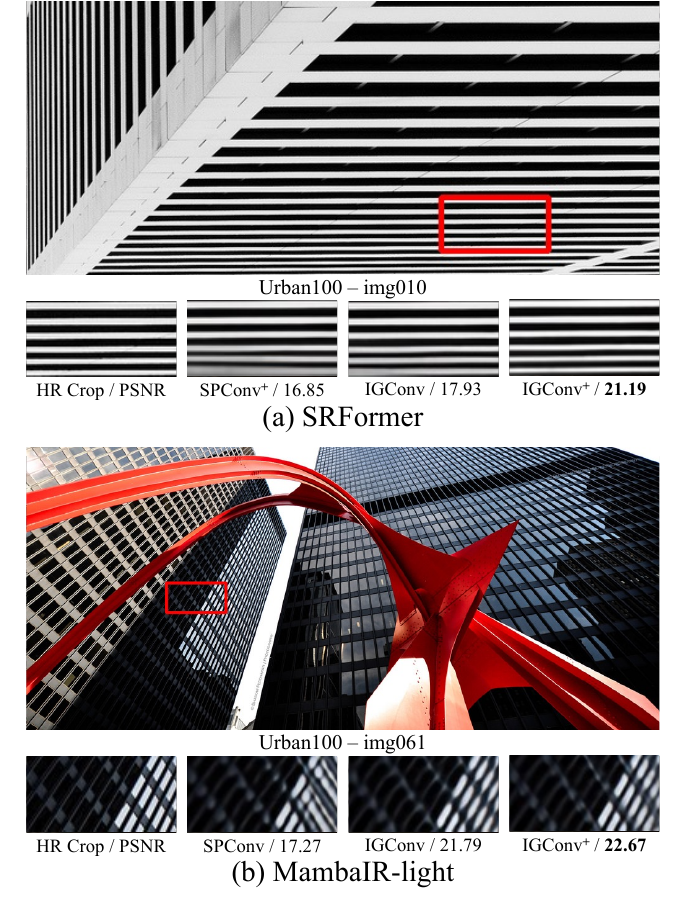}
  \vspace{-0.8cm}
  \caption{
    Visual comparisons on SPConv, SPConv$^{+}$, IGConv, and IGConv$^+$ on Urban100$\times$4 dataset. The best results on PSNR are bolded.
  }
  \label{fig:visual_spconv}
  \vspace{-0.5cm}
\end{figure}

\vspace{-0.1cm}
\subsection{Visual Results}
\vspace{-0.1cm}
To demonstrate that IGConv and IGConv$^{+}$ are also visually superior, we compare our methods visually to SPConv and then to SPConv$^{+}$.
As shown in Figure~\ref{fig:visual_spconv}, using IGConv and IGConv$^{+}$ improves the visual quality and the PSNR. 
These results demonstrate that our method yields visually pleasing results, emphasizing the potential of learning multi-scale and superior performance of our proposal.
\vspace{-0.1cm}
\section{Conclusion}
\vspace{-0.1cm}
This paper highlighted the inefficiency of the classic fixed-scale SR approach, which employs a scale-specific model for each targeted scale.
Based on the observation that encoder features are similar across scales and that SPConv operates in a highly correlated manner, we propose a multi-scale framework that employs a single encoder along with the IGConv.
Our multi-scale framework with IGConv significantly reduces both the training budget and parameter storage, achieving consistent performance across various encoders, regardless of its core operator, size, and training dataset.
Moreover, we introduced IGConv$^{+}$, which boosts performance by employing frequency loss and introducing IGSample, and FGRep. 
As a result, our ATD-IGConv$^{+}$ achieved a remarkable 0.21 dB improvement in PSNR on Urban100$\times4$ also reducing the training budget and stored parameters compared to the existing ATD.

\noindent\textbf{Discussion:} Multi-scale training with IGConv$^{+}$ significantly improved performance, but degree of improvement varies. 
These variations suggest that a new approach should be considered when proposing architectures suitable for multi-scale training and raise the need for further research. 

{
    \small
    \bibliographystyle{ieeenat_fullname}
    \bibliography{main}

\begin{thebibliography}{45}
\providecommand{\natexlab}[1]{#1}
\providecommand{\url}[1]{\texttt{#1}}
\expandafter\ifx\csname urlstyle\endcsname\relax
  \providecommand{\doi}[1]{doi: #1}\else
  \providecommand{\doi}{doi: \begingroup \urlstyle{rm}\Url}\fi

\bibitem[Agustsson and Timofte(2017)]{DIV2K}
Eirikur Agustsson and Radu Timofte.
\newblock Ntire 2017 challenge on single image super-resolution: Dataset and study.
\newblock In \emph{CVPRW}, 2017.

\bibitem[Bevilacqua et~al.(2012)Bevilacqua, Roumy, Guillemot, and Alberi-Morel]{Set5}
Marco Bevilacqua, Aline Roumy, Christine Guillemot, and Marie~Line Alberi-Morel.
\newblock Low-complexity single-image super-resolution based on nonnegative neighbor embedding.
\newblock 2012.

\bibitem[Cao et~al.(2023)Cao, Wang, Xian, Li, Ni, Pi, Zhang, Zhang, Timofte, and Van~Gool]{CiaoSR}
Jiezhang Cao, Qin Wang, Yongqin Xian, Yawei Li, Bingbing Ni, Zhiming Pi, Kai Zhang, Yulun Zhang, Radu Timofte, and Luc Van~Gool.
\newblock Ciaosr: Continuous implicit attention-in-attention network for arbitrary-scale image super-resolution.
\newblock In \emph{CVPR}, pages 1796--1807, 2023.

\bibitem[Chen et~al.(2023{\natexlab{a}})Chen, Xu, Hong, Tsai, Kuo, and Lee]{CLIT}
Hao-Wei Chen, Yu-Syuan Xu, Min-Fong Hong, Yi-Min Tsai, Hsien-Kai Kuo, and Chun-Yi Lee.
\newblock Cascaded local implicit transformer for arbitrary-scale super-resolution.
\newblock In \emph{CVPR}, pages 18257--18267, 2023{\natexlab{a}}.

\bibitem[Chen et~al.(2023{\natexlab{b}})Chen, Wang, Zhou, Qiao, and Dong]{HAT}
Xiangyu Chen, Xintao Wang, Jiantao Zhou, Yu Qiao, and Chao Dong.
\newblock Activating more pixels in image super-resolution transformer.
\newblock In \emph{CVPR}, pages 22367--22377, 2023{\natexlab{b}}.

\bibitem[Chen et~al.(2021)Chen, Liu, and Wang]{LIIF}
Yinbo Chen, Sifei Liu, and Xiaolong Wang.
\newblock Learning continuous image representation with local implicit image function.
\newblock In \emph{CVPR}, pages 8628--8638, 2021.

\bibitem[Chen et~al.(2024)Chen, Zhang, Gu, Kong, and Yang]{RGT}
Zheng Chen, Yulun Zhang, Jinjin Gu, Linghe Kong, and Xiaokang Yang.
\newblock Recursive generalization transformer for image super-resolution.
\newblock In \emph{ICLR}, 2024.

\bibitem[Ding et~al.(2019)Ding, Guo, Ding, and Han]{ACNet}
Xiaohan Ding, Yuchen Guo, Guiguang Ding, and Jungong Han.
\newblock Acnet: Strengthening the kernel skeletons for powerful cnn via asymmetric convolution blocks.
\newblock In \emph{ICCV}, pages 1911--1920, 2019.

\bibitem[Dong et~al.(2015)Dong, Loy, He, and Tang]{SRCNN}
Chao Dong, Chen~Change Loy, Kaiming He, and Xiaoou Tang.
\newblock Image super-resolution using deep convolutional networks.
\newblock \emph{IEEE TPAMI}, 38\penalty0 (2):\penalty0 295--307, 2015.

\bibitem[Dosovitskiy et~al.(2021)Dosovitskiy, Beyer, Kolesnikov, Weissenborn, Zhai, Unterthiner, Dehghani, Minderer, Heigold, Gelly, Uszkoreit, and Houlsby]{ViT}
Alexey Dosovitskiy, Lucas Beyer, Alexander Kolesnikov, Dirk Weissenborn, Xiaohua Zhai, Thomas Unterthiner, Mostafa Dehghani, Matthias Minderer, Georg Heigold, Sylvain Gelly, Jakob Uszkoreit, and Neil Houlsby.
\newblock An image is worth 16x16 words: Transformers for image recognition at scale.
\newblock \emph{ICLR}, 2021.

\bibitem[Gu and Dao(2023)]{Mamba}
Albert Gu and Tri Dao.
\newblock Mamba: Linear-time sequence modeling with selective state spaces.
\newblock \emph{arXiv preprint}, 2023.

\bibitem[Guo et~al.(2024)Guo, Li, Dai, Ouyang, Ren, and Xia]{MambaIR}
Hang Guo, Jinmin Li, Tao Dai, Zhihao Ouyang, Xudong Ren, and Shu-Tao Xia.
\newblock Mambair: A simple baseline for image restoration with state-space model.
\newblock In \emph{ECCV}, 2024.

\bibitem[He and Jin(2024)]{LMF}
Zongyao He and Zhi Jin.
\newblock Latent modulated function for computational optimal continuous image representation.
\newblock In \emph{CVPR}, pages 26026--26035, 2024.

\bibitem[Hu et~al.(2019)Hu, Mu, Zhang, Wang, Tan, and Sun]{MetaSR}
Xuecai Hu, Haoyuan Mu, Xiangyu Zhang, Zilei Wang, Tieniu Tan, and Jian Sun.
\newblock Meta-sr: A magnification-arbitrary network for super-resolution.
\newblock In \emph{CVPR}, pages 1575--1584, 2019.

\bibitem[Huang et~al.(2015)Huang, Singh, and Ahuja]{Urban100}
Jia-Bin Huang, Abhishek Singh, and Narendra Ahuja.
\newblock Single image super-resolution from transformed self-exemplars.
\newblock In \emph{CVPR}, pages 5197--5206, 2015.

\bibitem[Kim et~al.(2016)Kim, Lee, and Lee]{VDSR}
Jiwon Kim, Jung~Kwon Lee, and Kyoung~Mu Lee.
\newblock Accurate image super-resolution using very deep convolutional networks.
\newblock In \emph{CVPR}, pages 1646--1654, 2016.

\bibitem[Kornblith et~al.(2019)Kornblith, Norouzi, Lee, and Hinton]{CKA}
Simon Kornblith, Mohammad Norouzi, Honglak Lee, and Geoffrey Hinton.
\newblock Similarity of neural network representations revisited.
\newblock In \emph{ICML}, pages 3519--3529. PMLR, 2019.

\bibitem[Lai et~al.(2017)Lai, Huang, Ahuja, and Yang]{LapSRN}
Wei-Sheng Lai, Jia-Bin Huang, Narendra Ahuja, and Ming-Hsuan Yang.
\newblock Deep laplacian pyramid networks for fast and accurate super-resolution.
\newblock In \emph{CVPR}, pages 624--632, 2017.

\bibitem[Lee and Jin(2022)]{LTE}
Jaewon Lee and Kyong~Hwan Jin.
\newblock Local texture estimator for implicit representation function.
\newblock In \emph{CVPR}, pages 1929--1938, 2022.

\bibitem[Li et~al.(2023)Li, Zhang, Liu, and Zhu]{CRAFT}
Ao Li, Le Zhang, Yun Liu, and Ce Zhu.
\newblock Feature modulation transformer: Cross-refinement of global representation via high-frequency prior for image super-resolution.
\newblock In \emph{ICCV}, pages 12514--12524, 2023.

\bibitem[Liang et~al.(2021)Liang, Cao, Sun, Zhang, Van~Gool, and Timofte]{SwinIR}
Jingyun Liang, Jiezhang Cao, Guolei Sun, Kai Zhang, Luc Van~Gool, and Radu Timofte.
\newblock Swinir: Image restoration using swin transformer.
\newblock In \emph{CVPRW}, pages 1833--1844, 2021.

\bibitem[Lim et~al.(2017)Lim, Son, Kim, Nah, and Mu~Lee]{EDSR}
Bee Lim, Sanghyun Son, Heewon Kim, Seungjun Nah, and Kyoung Mu~Lee.
\newblock Enhanced deep residual networks for single image super-resolution.
\newblock In \emph{CVPRW}, pages 136--144, 2017.

\bibitem[Liu et~al.(2023{\natexlab{a}})Liu, Lu, Fu, and Cao]{DySample}
Wenze Liu, Hao Lu, Hongtao Fu, and Zhiguo Cao.
\newblock Learning to upsample by learning to sample.
\newblock In \emph{ICCV}, pages 6027--6037, 2023{\natexlab{a}}.

\bibitem[Liu et~al.(2023{\natexlab{b}})Liu, Dong, Liang, Liu, Dong, Chen, Chen, Fu, and Wang]{DITN}
Yong Liu, Hang Dong, Boyang Liang, Songwei Liu, Qingji Dong, Kai Chen, Fangmin Chen, Lean Fu, and Fei Wang.
\newblock Unfolding once is enough: A deployment-friendly transformer unit for super-resolution.
\newblock In \emph{ACMMM}, pages 7952--7960, 2023{\natexlab{b}}.

\bibitem[Martin et~al.(2001)Martin, Fowlkes, Tal, and Malik]{BSD100}
David Martin, Charless Fowlkes, Doron Tal, and Jitendra Malik.
\newblock A database of human segmented natural images and its application to evaluating segmentation algorithms and measuring ecological statistics.
\newblock In \emph{ICCV}, pages 416--423. IEEE, 2001.

\bibitem[Matsui et~al.(2017)Matsui, Ito, Aramaki, Fujimoto, Ogawa, Yamasaki, and Aizawa]{Manga109}
Yusuke Matsui, Kota Ito, Yuji Aramaki, Azuma Fujimoto, Toru Ogawa, Toshihiko Yamasaki, and Kiyoharu Aizawa.
\newblock Sketch-based manga retrieval using manga109 dataset.
\newblock \emph{Multimedia tools and applications}, 76:\penalty0 21811--21838, 2017.

\bibitem[Mildenhall et~al.(2020)Mildenhall, Srinivasan, Tancik, Barron, Ramamoorthi, and Ng]{NERF}
Ben Mildenhall, Pratul~P. Srinivasan, Matthew Tancik, Jonathan~T. Barron, Ravi Ramamoorthi, and Ren Ng.
\newblock Nerf: Representing scenes as neural radiance fields for view synthesis.
\newblock In \emph{ECCV}, 2020.

\bibitem[Ray et~al.(2024)Ray, Kumar, and Kolekar]{CFAT}
Abhisek Ray, Gaurav Kumar, and Maheshkumar~H Kolekar.
\newblock Cfat: Unleashing triangular windows for image super-resolution.
\newblock In \emph{CVPR}, pages 26120--26129, 2024.

\bibitem[Shi et~al.(2016)Shi, Caballero, Husz{\'a}r, Totz, Aitken, Bishop, Rueckert, and Wang]{ESPCN}
Wenzhe Shi, Jose Caballero, Ferenc Husz{\'a}r, Johannes Totz, Andrew~P Aitken, Rob Bishop, Daniel Rueckert, and Zehan Wang.
\newblock Real-time single image and video super-resolution using an efficient sub-pixel convolutional neural network.
\newblock In \emph{CVPR}, pages 1874--1883, 2016.

\bibitem[Song et~al.(2023)Song, Sun, Zhang, Su, Shi, and He]{OPESR}
Gaochao Song, Qian Sun, Luo Zhang, Ran Su, Jianfeng Shi, and Ying He.
\newblock Ope-sr: Orthogonal position encoding for designing a parameter-free upsampling module in arbitrary-scale image super-resolution.
\newblock In \emph{CVPR}, pages 10009--10020, 2023.

\bibitem[Sun et~al.(2022)Sun, Pan, and Tang]{ShuffleMixer}
Long Sun, Jinshan Pan, and Jinhui Tang.
\newblock Shufflemixer: An efficient convnet for image super-resolution.
\newblock \emph{NeurIPS}, 35:\penalty0 17314--17326, 2022.

\bibitem[Timofte et~al.(2017)Timofte, Agustsson, Van~Gool, Yang, and Zhang]{DF2KDataset}
Radu Timofte, Eirikur Agustsson, Luc Van~Gool, Ming-Hsuan Yang, and Lei Zhang.
\newblock Ntire 2017 challenge on single image super-resolution: Methods and results.
\newblock In \emph{CVPRW}, pages 114--125, 2017.

\bibitem[Tu et~al.(2022)Tu, Talebi, Zhang, Yang, Milanfar, Bovik, and Li]{MAXIM}
Zhengzhong Tu, Hossein Talebi, Han Zhang, Feng Yang, Peyman Milanfar, Alan Bovik, and Yinxiao Li.
\newblock Maxim: Multi-axis mlp for image processing.
\newblock In \emph{CVPR}, pages 5769--5780, 2022.

\bibitem[Vasconcelos et~al.(2023)Vasconcelos, Oztireli, Matthews, Hashemi, Swersky, and Tagliasacchi]{CUF}
Cristina~N Vasconcelos, Cengiz Oztireli, Mark Matthews, Milad Hashemi, Kevin Swersky, and Andrea Tagliasacchi.
\newblock Cuf: Continuous upsampling filters.
\newblock In \emph{CVPR}, pages 9999--10008, 2023.

\bibitem[Vaswani et~al.(2017)Vaswani, Shazeer, Parmar, Uszkoreit, Jones, Gomez, Kaiser, and Polosukhin]{Transformers}
Ashish Vaswani, Noam Shazeer, Niki Parmar, Jakob Uszkoreit, Llion Jones, Aidan~N Gomez, {\L}ukasz Kaiser, and Illia Polosukhin.
\newblock Attention is all you need.
\newblock \emph{NeurIPS}, 30, 2017.

\bibitem[Wang et~al.(2023)Wang, Chen, Ni, Wang, Tong, and Liu]{EQSR}
Xiaohang Wang, Xuanhong Chen, Bingbing Ni, Hang Wang, Zhengyan Tong, and Yutian Liu.
\newblock Deep arbitrary-scale image super-resolution via scale-equivariance pursuit.
\newblock In \emph{CVPR}, pages 1786--1795, 2023.

\bibitem[Zamir et~al.(2022)Zamir, Arora, Khan, Hayat, Khan, and Yang]{Restormer}
Syed~Waqas Zamir, Aditya Arora, Salman Khan, Munawar Hayat, Fahad~Shahbaz Khan, and Ming-Hsuan Yang.
\newblock Restormer: Efficient transformer for high-resolution image restoration.
\newblock In \emph{CVPR}, pages 5728--5739, 2022.

\bibitem[Zeyde et~al.(2012)Zeyde, Elad, and Protter]{Set14}
Roman Zeyde, Michael Elad, and Matan Protter.
\newblock On single image scale-up using sparse-representations.
\newblock In \emph{Curves and Surfaces: 7th International Conference, Avignon, France, June 24-30, 2010, Revised Selected Papers 7}, pages 711--730. Springer, 2012.

\bibitem[Zhang et~al.(2024{\natexlab{a}})Zhang, Li, Zhou, Zhao, and Gu]{ATD}
Leheng Zhang, Yawei Li, Xingyu Zhou, Xiaorui Zhao, and Shuhang Gu.
\newblock Transcending the limit of local window: Advanced super-resolution transformer with adaptive token dictionary.
\newblock In \emph{CVPR}, pages 2856--2865, 2024{\natexlab{a}}.

\bibitem[Zhang et~al.(2022)Zhang, Zeng, Guo, and Zhang]{ELAN}
Xindong Zhang, Hui Zeng, Shi Guo, and Lei Zhang.
\newblock Efficient long-range attention network for image super-resolution.
\newblock In \emph{ECCV}, pages 649--667. Springer, 2022.

\bibitem[Zhang et~al.(2024{\natexlab{b}})Zhang, Zhang, and Yu]{HiTSR}
Xiang Zhang, Yulun Zhang, and Fisher Yu.
\newblock Hit-sr: Hierarchical transformer for efficient image super-resolution.
\newblock In \emph{ECCV}, 2024{\natexlab{b}}.

\bibitem[Zhang et~al.(2018{\natexlab{a}})Zhang, Li, Li, Wang, Zhong, and Fu]{RCAN}
Yulun Zhang, Kunpeng Li, Kai Li, Lichen Wang, Bineng Zhong, and Yun Fu.
\newblock Image super-resolution using very deep residual channel attention networks.
\newblock In \emph{ECCV}, pages 286--301, 2018{\natexlab{a}}.

\bibitem[Zhang et~al.(2018{\natexlab{b}})Zhang, Tian, Kong, Zhong, and Fu]{RDN}
Yulun Zhang, Yapeng Tian, Yu Kong, Bineng Zhong, and Yun Fu.
\newblock Residual dense network for image super-resolution.
\newblock In \emph{CVPR}, pages 2472--2481, 2018{\natexlab{b}}.

\bibitem[Zheng et~al.(2024)Zheng, Sun, Dong, and Pan]{SMFANet}
Mingjun Zheng, Long Sun, Jiangxin Dong, and Jinshan Pan.
\newblock Smfanet: A lightweight self-modulation feature aggregation network for efficient image super-resolution.
\newblock In \emph{ECCV}, 2024.

\bibitem[Zhou et~al.(2023)Zhou, Li, Guo, Bai, Cheng, and Hou]{SRFormer}
Yupeng Zhou, Zhen Li, Chun-Le Guo, Song Bai, Ming-Ming Cheng, and Qibin Hou.
\newblock Srformer: Permuted self-attention for single image super-resolution.
\newblock In \emph{ICCV}, pages 12780--12791, 2023.

\end{thebibliography}
}

\clearpage
\setcounter{page}{1}
\maketitlesupplementary

The supplementary includes reasons not considering pre-upsampling architecture, detailed comparisons with similar methods, training details, ablation studies, experiments beyond the $\times$4 scale, additional visualizations of inter-scale correlations, quantitative and qualitative results compared to Arbitrary-Scale Super-Resolution~(ASSR) methods, and finally, visual results on Out-Of-Distribution~(OOD) scales.

\vspace{-0.1cm}
\section{Pre-Upsampling Architecture}
\vspace{-0.1cm}
Recently, most methods employing neural networks have adopted a post-upsampling architecture that extracts low-resolution~(LR) features using an encoder and upsamples them in the final step to achieve super-resolution~(SR).
However, some early studies~\cite{SRCNN, VDSR} employed a pre-upsampling architecture, where LR images are upsampled using bicubic interpolation, followed by post-processing through a neural network to achieve SR.
Pre-upsampling architecture can easily predict any arbitrary scale, but it requires tremendous computations since all operations are performed in high-resolution~(HR) space.
For this reason, recent studies targeting only SR do not consider pre-upsampling architecture, and we have also not mentioned it in the main manuscript.

\vspace{-0.1cm}
\section{Comparions on LapSRN and MDSR}\label{sec:multiscale_comparison}
\vspace{-0.1cm}
Before our research, attempts were made to train multiple scales simultaneously employing a single model.
For example, LapSRN~\cite{LapSRN} aimed to stably train and predict $\times$8 scale by progressively upsampling the LR image. 
MDSR~\cite{EDSR} extracts features from images and converts them to RGB images using scale-specific heads and tails while sharing a single feature extractor across all scales~($\times$2, $\times$3, and $\times$4) to refine features from multiple scales. 
Our approach offers several advantages over LapSRN and MDSR.
First, LapSRN requires computations in the HR space because of its progressive upsampling design, resulting in a significant computational burden.
We demonstrate in Appendix~\ref{sec:scale8} that our method can predict the $\times$8 scale without such excessive computational costs brought by progressive upscaling architecture.
Additionally, MDSR independently trains each head and tail, which means heads and tails cannot learn multi-scale information, potentially negatively impacting performance. 

\vspace{-0.1cm}
\section{Comparisons on CUF}
\vspace{-0.1cm}
The continuous upsampling filters (CUF)~\cite{CUF} is similar to our method in that it maps scale-equivariant conditions into convolution kernels using an INR-based hyper-network and is converted efficiently when instantiated at the specific integer scales. 
However, our method offers some advantages in computational efficiency.
To demonstrate our efficiency compared to CUF, we consider only the instantiated CUF, excluding the inefficiencies that arise from targeting Arbitrary-Scale Super-Resolution (ASSR).
CUF performs scale-specific modulation using depth-wise convolution followed by depth-to-space~($\mathcal{DS}$) to upsample, with two additional point-wise convolutions added to compensate for insufficient channel mixing.
These additional point-wise convolutions in HR space result in significant computational overhead, similar to SPConv$^{+}$. 
In Appendix~\ref{sec:assr}, we demonstrate that instantiated IGConv$^{+}$ outperforms instantiated CUF in both efficiency and performance, highlighting the superiority of our approach that performs all heavy computation in LR space.

\vspace{-0.1cm}
\section{Training Details}\label{sec:training_details}
\vspace{-0.1cm}
This section describes the training details for each method.
The training details are presented in Table~\ref{tab:trainingdetails}.
The training budget allocated to our framework matches that required by scale-specific methods for training at the $\times$2 scale. 
Consequently, our framework achieves a one-third reduction in the overall training budget (in terms of training time or GPU usage) compared to fixed-scale methods utilizing SPConv or SPConv$^{+}$.

\begin{table}[!h]
\caption{
    Training details for each method.
}\label{tab:trainingdetails}
\vspace{-0.2cm}
\resizebox{\columnwidth}{!}{%
\begin{tabular}{l|ccccc}
Methods & PatchSize & BatchSize & Iteration & LR & EMA \\ \hline
EDSR~\cite{EDSR} & 48 & 16 & 300000 & 0.0001 & \CheckmarkBold \\
RCAN~\cite{RCAN} & 48 & 16 & 1000000 & 0.0001 & \XSolidBrush \\
SMFANet$+$~\cite{SMFANet} & 64 & 64 & 1000000 & 0.001 & \CheckmarkBold \\
HiT-SRF~\cite{HiTSR} & 64 & 64 & 500000 & 0.0005 & \XSolidBrush \\
SRFormer~\cite{SRFormer} & 64 & 32 & 500000 & 0.0002 & \XSolidBrush \\
MambaIR-light~\cite{MambaIR} & 64 & 32 & 500000 & 0.0002 & \XSolidBrush \\
MambaIR~\cite{MambaIR} & 64 & 32 & 500000 & 0.0002 & \XSolidBrush \\ \hline
\end{tabular}%
\vspace{-0.5cm}
}
\end{table}

The training details of the methods adopting pre-training and fine-tuning strategies can be found in Table~\ref{tab:pretrain_trainingdetails}. 
ATD~\cite{ATD} adopted the pre-training strategy with a small patch size on the DF2K dataset, followed by fine-tuning with a larger patch size. 
In contrast, HAT~\cite{HAT} is pre-trained on the more extensive ImageNet dataset before being fine-tuned on the DF2K dataset.

\begin{table}[!h]
\caption{Training details for methods adopting the pre-training strategy.}\label{tab:pretrain_trainingdetails}
\vspace{-0.2cm}
\resizebox{\columnwidth}{!}{%
\begin{tabular}{l|l|cccc}
Methods & (Phase)~Dataset & PatchSize & BatchSize & Iteration & LR \\ \hline
\multirow{2}{*}{ATD~\cite{ATD}} & (1)~DF2K & 64 & 32 & 300K & 0.0002 \\
 & (2)~DF2K & 96 & 32 & 250K & 0.0002 \\ \hline
\multirow{2}{*}{HAT~\cite{HAT}} & (1)~ImageNet & 64 & 32 & 800K & 0.0002 \\
 & (2)~DF2K & 64 & 32 & 250K & 0.00001 \\ \hline
\end{tabular}%
}
\vspace{-0.5cm}
\end{table}

\begin{table*}[!htb]
\caption{
    Comparisons of fixed-scale upsamplers~(SPConv$^{+}$) and our proposed multi-scale upsamplers~(IGConv$^{+}$) on RCAN encoder at 4 scales~($\times$2, $\times3$, $\times$4, and $\times$8). Results from SPConv$^{+}${\scriptsize($\times r$)} are measured by each scale-specific model, while results from IGConv and IGConv$^{+}$ are measured by \textbf{a single model}.
}\label{tab:scale8}
\resizebox{\textwidth}{!}{%
\begin{tabular}{l|c|l|ccccc}
Encoder & Scale & Upsampler & Set5 & Set14 & B100 & Urban100 & Manga109 \\ \hline
\multirow{8}{*}{RCAN~\cite{RCAN}} & \multirow{2}{*}{2} & SPConv$^{+}${\scriptsize($\times$2)} & 38.27/0.9614 & 34.12/0.9216 & 32.41/0.9027 & 33.34/0.9384 & 39.44/0.9786 \\
 &  & IGConv$^{+}$ & 38.23/0.9614 & 34.12/0.9217 & 32.38/0.9022 & 33.27/0.9383 & 39.39/0.9784 \\ \cline{2-8} 
 & \multirow{2}{*}{3} & SPConv$^{+}${\scriptsize($\times$3)} & 34.74/0.9299 & 30.65/0.8482 & 29.32/0.8111 & 29.09/0.8702 & 34.44/0.9499 \\
 &  & IGConv$^{+}$ & 34.86/0.9306 & 30.71/0.8490 & 29.34/0.8111 & 29.18/0.8712 & 34.66/0.9505 \\ \cline{2-8} 
 & \multirow{2}{*}{4} & SPConv$^{+}${\scriptsize($\times$4)} & 32.63/0.9002 & 28.87/0.7889 & 27.77/0.7436 & 26.82/0.8087 & 31.22/0.9173 \\
 &  & IGConv$^{+}$ & 32.68/0.9007 & 28.98/0.7907 & 27.83/0.7444 & 27.03/0.8118 & 31.60/0.9182 \\ \cline{2-8} 
 & \multirow{2}{*}{8} & SPConv$^{+}${\scriptsize($\times$8)} & 27.31/0.7878 & 25.23/0.6511 & 24.98/0.6058 & 23.00/0.6452 & 25.24/0.8029 \\
 &  & IGConv$^{+}$ & 27.34/0.7850 & 25.35/0.6513 & 25.04/0.6048 & 23.13/0.6454 & 25.45/0.8016 \\ \hline
\end{tabular}%
}
\end{table*}

\section{Beyond $\times$4 Scale}\label{sec:scale8}
Many recent SR studies have focused on three scales ($\times$2, $\times$3, and $\times$4), and accordingly, we also trained and evaluated our framework on these three scales. 
However, some previous studies have evaluated four scales including $\times$8.
We wonder whether more challenging and complex scales can also be trained simultaneously.
For this reason, we use RCAN~\cite{RCAN} as the encoder to train and evaluate our model on four scales simultaneously including $\times$8. 
As shown in Table~\ref{tab:scale8}, IGConv$^{+}$ achieves 0.13 dB improvements on PSNR on Urban100$\times$8 compared to SPConv$^{+}${\scriptsize($\times$8)}.
Note that IGConv$^{+}$ is trained for four scales simultaneously, compared to existing RCAN with SPConv$^{+}$.
This surprising result underscores our claim that a fixed-scale training approach is unnecessary and shows that learning higher scales is possible without a progressive upsampling architecture. 

\section{Ablation Study}\label{sec:ablation}
To validate that every proposed method contributes to performance improvement, we conduct an ablation study by adding our proposal to RDN~\cite{RDN}.
Table~\ref{tab:ablation} demonstrates the performance improvement by replacing the upsampler with IGConv and adding frequency loss, IGSample, and FGRep.
This indicates that our proposed methods effectively contribute to performance improvement.

\begin{table}[!h]
\caption{
    Ablation study on our proposed methods. FFT, IGS, and FGR denote frequency loss, IGSample, and FGRep, respectively. 
}\label{tab:ablation}
\resizebox{\columnwidth}{!}{
\begin{tabular}{c|ccc|ccc}
\multirow{2}{*}{Upsampler} & \multirow{2}{*}{FFT} & \multirow{2}{*}{IGS} & \multirow{2}{*}{FGR} & \multicolumn{3}{c}{Urban100~(PSNR)} \\ \cline{5-7} 
 % &  &  &  & \multicolumn{3}{c|}{Seen} \\ %& \multicolumn{2}{c}{Unseen} 
 &  &  &  & $\times$2 & $\times$3 & \multicolumn{1}{c}{$\times$4}  \\ \hline %& $\times$6 & $\times$8 \\ \hline
 SPConv$^{+}$ &   &   &   & 32.89 & 28.80 & 26.61 \\ \cline{1-1} % & - & - \\ \cline{1-1}
IGConv &   &   &   & 33.06 & 28.97 & 26.82 \\ \cline{1-1} % & 24.13 & 22.76 \\ \cline{1-1}
\multirow{3}{*}{IGConv$^+$} & \CheckmarkBold &   &   & 33.10  & 29.02 & 26.88 \\ % & 24.17 & 22.78 \\ 
 & \CheckmarkBold & \CheckmarkBold &   & 33.15 & 29.08 & 26.95 \\ % & 24.24 & 22.84 \\ 
 & \CheckmarkBold & \CheckmarkBold & \CheckmarkBold & 33.17 & 29.12 & 26.96 \\ \hline % & 24.25 & 22.84 \\ \hline
\end{tabular}
}
\end{table}

\section{More Visualizations on Inter-Scale Corr.}\label{sec:igvis_full}
\begin{figure}[h]
  \centering
  \includegraphics[width=\columnwidth]{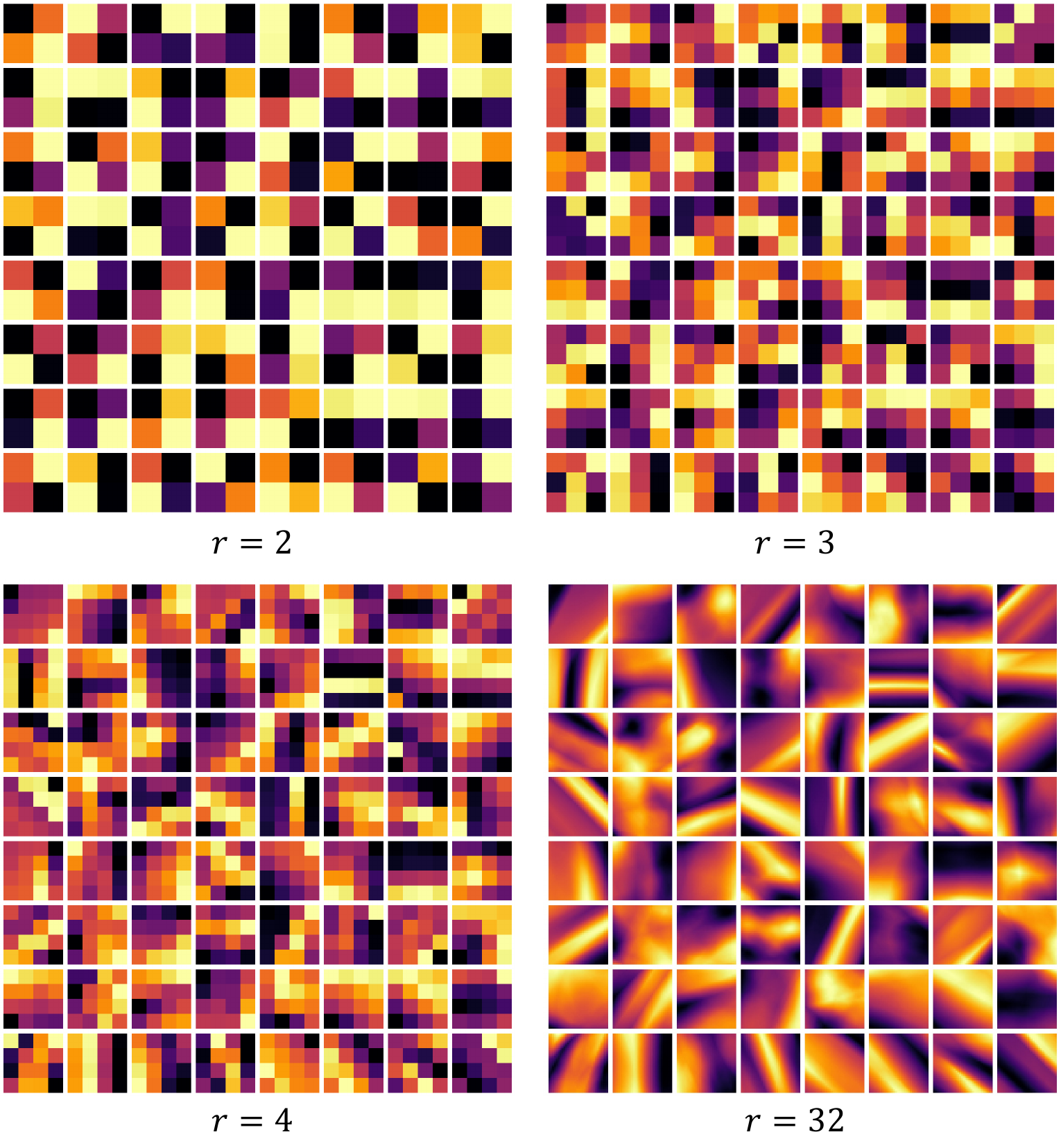}
  \caption{
    Visualizations of convolution filters inferred by $\mathcal{H}$ of RDN-IGConv$^{+}$ for scales $\times$2, $\times$3, $\times$4, and $\times$32. 
  }
  \label{fig:implicitgridsfull}
\end{figure}

This section includes additional visualizations of convolution filters trained for capturing inter-scale correlations. 
As shown in Figure~\ref{fig:implicitgridsfull}, all convolution filters from $\mathcal{H}$ of RDN-IGConv$^{+}$ change continuously with variations in scale, demonstrating that $\mathcal{H}$ effectively captures the inter-scale correlation.

\begin{table*}[h]
\caption{
    Comparison of ASSR upsamplers with our proposal. Efficiency measures are calculated by upsampling a 128$\times$128 image using an A6000 GPU for $\times$4 scale. The best and second-best results are highlighted in bold and underlined, respectively.
    $^{\S}$ denotes each method is instantiated to the $\times$4 scale.
}
\resizebox{\textwidth}{!}{%
\begin{tabular}{ll|ccc|ccc|ccc|ccc|ccc}
\multirow{2}{*}{Encoder} & \multirow{2}{*}{Upsampler} & \multirow{2}{*}{\begin{tabular}[c]{@{}c@{}}Latency\\ (ms)\end{tabular}} & \multirow{2}{*}{\begin{tabular}[c]{@{}c@{}}\#Params\\ (K)\end{tabular}} & \multirow{2}{*}{\begin{tabular}[c]{@{}c@{}}Memory \\ (mb)\end{tabular}} & \multicolumn{3}{c}{Set5} & \multicolumn{3}{c}{Set14} & \multicolumn{3}{c}{B100} & \multicolumn{3}{c}{Urban100} \\
 &  &  &  &  & $\times$2 & $\times$3 & $\times$4 & $\times$2 & $\times$3 & $\times$4 & $\times$2 & $\times$3 & $\times$4 & $\times$2 & $\times$3 & $\times$4 \\ \hline

% RDN~\cite{RDN} -------------------------------------------------------
\multirow{7}{*}{RDN~\cite{RDN}} & LIIF~\cite{LIIF} & 139.6 & 347 & 1811 & 38.17 & 34.68 & 32.50 & 33.97 & 30.53 & 28.80 & 32.32 & 29.26 & 27.74 & 32.87 & 28.82 & 26.68 \\
 & LTE~\cite{LTE}  & 166.0 & 494 & 1608 & 38.23 & 34.72 & 32.61 & 34.09 & 30.58 & 28.88 & 32.36 & 29.30 & 27.77 & 33.04 & 28.97 & 26.81 \\
 & CiaoSR~\cite{CiaoSR} & 395.5 & 1429 & 12378 & \textbf{38.29} & \textbf{34.85} & \textbf{32.66} & \textbf{34.22} & \underline{30.65} & \textbf{28.93} & \textbf{32.41} & \textbf{29.34} & \textbf{27.83} & \textbf{33.30} & \textbf{29.17} & \textbf{27.11} \\
 & LM-LTE~\cite{LMF} & 31.2 & 271 & 367 & 38.23 & \underline{34.76} & 32.53 & \underline{34.11} & 30.56 & 28.86 & 32.37 & 29.31 & 27.78 & 33.03 & 28.96 & 26.80 \\
 & OPE-SR~\cite{OPESR} & 15.6 & \textbf{0} & 339 & 37.60 & 34.59 & 32.47 & 33.39 & 30.49 & 28.80 & 32.05 & 29.19 & 27.72 & 31.78 & 28.63 & 26.53 \\
 & CUF~\cite{CUF} & - / 1.2$^{\S}$ & 10 & - / 132$^{\S}$  & 38.23 & 34.72 & 32.54 & 33.99 & 30.58 & 28.86 & 32.35 & 29.29 & 27.76 &  33.01 & 28.91 & 26.75 \\
 & \textbf{IGConv$^{+}$~(Ours)} & \textbf{2.3} / \textbf{0.5}$^{\S}$ & 922 & \textbf{71} / \textbf{43}$^{\S}$  & \underline{38.26} & 34.74 & \underline{32.64} & 34.10 & \textbf{30.68} & \underline{28.91} & \underline{32.39} & \underline{29.33} & \underline{27.82} & \underline{33.17} & \underline{29.11} & \underline{26.96} \\ \hline
 
% SwinIR~\cite{SwinIR} -------------------------------------------------------
 \multirow{6}{*}{SwinIR~\cite{SwinIR}} & LIIF~\cite{LIIF} & 342.8 & 614 & 5015 & 38.28 & 34.87 & 32.73 & 34.14 & 30.75 & 28.98 & 32.39 & 29.34 & 27.84 & 33.36 & 29.33 & 27.15 \\
 & LTE~\cite{LTE} & 166.0 & 1028 & 1619 & 38.33 & \underline{34.89} & \underline{32.81} & 34.25 & 30.80 & 29.06 & 32.44 & 29.39 & 27.86 & 33.50 & 29.41 & 27.24 \\
 & CiaoSR~\cite{CiaoSR} & 889.9 & 3168 & 34760 & \textbf{38.38} & \textbf{34.91} & \textbf{32.84} & \textbf{34.33} & \underline{30.82} & \underline{29.08} & \textbf{32.47} & \textbf{29.42} & \underline{27.90} & \textbf{33.65} & \underline{29.52} & \textbf{27.42} \\
 & LM-LTE~\cite{LMF} & 31.4 & 538 & 376 & 38.32 & 34.88 & 32.77 & 34.28 & 30.79 & 29.01 & \underline{32.46} & 29.39 & 27.87 & 33.52 & 29.44 & 27.24  \\ 
 & CUF~\cite{CUF} & - / 3.6$^{\S}$ & \textbf{37} & - / 376$^{\S}$ & 38.34 & 34.88 & 32.80 & \underline{34.29} & 30.79 & 29.02 & 32.45 & 29.38 & 27.85 & 33.54 & 29.45 & 27.24 \\
 & \textbf{IGConv$^{+}$~(Ours)} & \textbf{4.6} / \textbf{0.8}$^{\S}$ & 1991 & \textbf{215} / \textbf{52}$^{\S}$ & \underline{38.35} & \underline{34.89} & 32.79 & 34.18 & \textbf{30.84} & \textbf{29.09} & \underline{32.46} & \underline{29.41} & \textbf{27.91} & \underline{33.60} & \textbf{29.53} & \underline{27.35} \\ \hline
\end{tabular}%
}
% \vspace{-0.4cm}
\end{table*}

\begin{table*}[h]
\caption{
    Comparison of ASSR upsamplers with our proposal trained for arbitrary-scale on non-integer scales. The best and second-best results are highlighted in bold and underlined, respectively.
}\label{tab:sota_float}
\resizebox{\textwidth}{!}{%
\begin{tabular}{ll|ccc|ccc|ccc|ccc}
\multirow{2}{*}{Encoder} & \multirow{2}{*}{Upsampler} & \multicolumn{3}{c|}{Set5} & \multicolumn{3}{c|}{Set14} & \multicolumn{3}{c|}{B100} & \multicolumn{3}{c}{Urban100} \\
 &  & $\times$1.5 & $\times$2.5 & $\times$3.5 & $\times$1.5 & $\times$2.5 & $\times$3.5 & $\times$1.5 & $\times$2.5 & $\times$3.5 & $\times$1.5 & $\times$2.5 & $\times$3.5 \\ \hline
\multirow{5}{*}{RDN~\cite{RDN}} & LIIF~\cite{LIIF} & 41.43 & 36.15 & 33.56 & 37.45 & 31.87 & 29.56 & 35.83 & 30.48 & 28.42 & \underline{36.79} & 30.49 & 27.64 \\
 & LTE~\cite{LTE} & \textbf{41.51} & \underline{36.18} & \textbf{33.64} & \textbf{37.55} & \underline{31.91} & \underline{29.62} & \underline{35.87} & \underline{30.51} & \underline{28.45} & \textbf{36.97} & \underline{30.64} & \underline{27.77} \\
 & LM-LTE~\cite{LMF} & \underline{41.49} & \underline{36.18} & \underline{33.62} & \underline{37.52} & \underline{31.91} & 29.58 & \textbf{35.88} & \textbf{30.52} & \underline{28.45} & \textbf{36.97} & 30.62 & \underline{27.77} \\
 & OPE-SR~\cite{OPESR} & 40.24 & 35.85 & 33.49 & 36.02 & 31.67 & 29.55 & 35.07 & 30.33 & 28.38 & 33.47 & 30.08 & 27.48 \\
 & \textbf{IGConv$^{+}_{arb}$} & 41.25 & \textbf{36.20} & \textbf{33.64} & 37.40 & \textbf{32.00} & \textbf{29.68} & 35.74 & \underline{30.51} & \textbf{28.46} & 36.64 & \textbf{30.70} & \textbf{27.85} \\ \hline
\end{tabular}
}
\vspace{-0.4cm}
\end{table*}

\section{Comparisons on Arbitrary-Scale Methods}\label{sec:assr}
We also compare our method to ASSR methods since they share similar architecture~(a pair of a single encoder and single upsampler).
For the comparison, we train and evaluate our upsampler, IGConv$^{+}$, using RDN~\cite{RDN} and SwinIR~\cite{SwinIR}, which are commonly used as encoders in ASSR methods. 
The baselines for comparison include LIIF~\cite{LIIF}, LTE~\cite{LTE}, CiaoSR~\cite{CiaoSR}, OPE-SR~\cite{OPESR}, CUF~\cite{CUF}, and LM-LTE~\cite{LMF}.
As shown in Table~\ref{tab:sota_float}, SwinIR-IGConv$^{+}$ achieves a 0.01 dB higher PSNR on B100$\times$4 compared to SwinIR-CiaoSR while reducing latency and memory usage by 99.5\% and 99.4\%, respectively, demonstrating exceptional performance-efficiency trade-off.
Moreover, instantiated IGConv$^{+}$ outperforms the instantiated CUF achieving a 0.06 dB higher PSNR on Urban100$\times$2, with 78\% and 86\% lower latency and memory usage, respectively, suggesting that our efficiency is not merely due to the absence of non-integer scale prediction.

\begin{figure}[h!]
  \centering
  \includegraphics[width=\columnwidth]{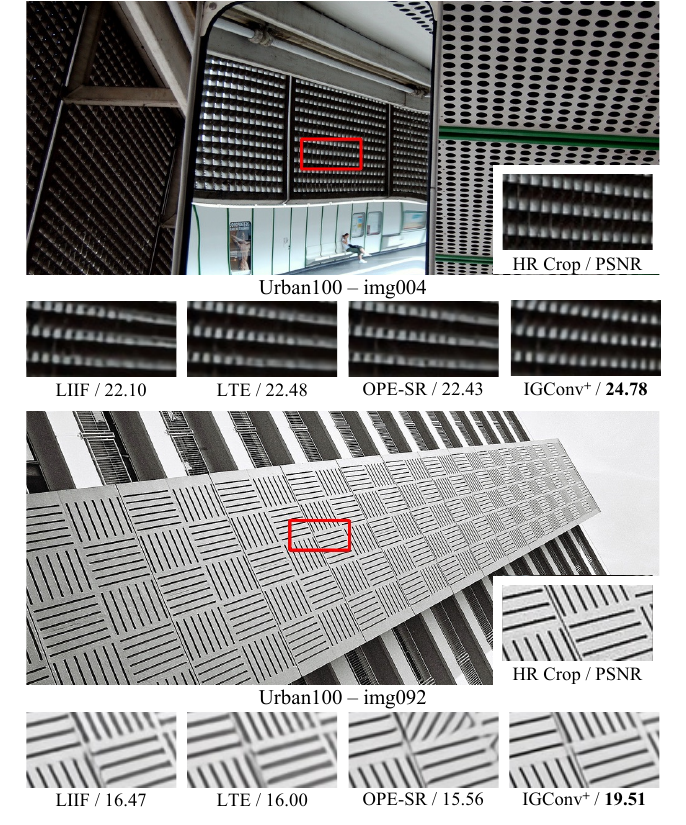}
  \caption{
    Visual comparisons on IGConv$^{+}$ and ASSR methods using RDN~\cite{RDN} encoder on Urban100$\times$4 dataset.
    The best result on PSNR is bolded.
  }
  \label{fig:visual_assr}
\end{figure}

We also visually compare IGConv$^{+}$ with the ASSR methods~(LIIF, LTE, and OPE-SR).  
In Figure.~\ref{fig:visual_assr}, IGConv$^{+}$ demonstrates better visual quality and PSNR compared to the ASSR method.
This confirms that IGConv$^{+}$ is efficient and has superior visual quality to the ASSR methods, highlighting the exceptional performance-efficiency trade-off of our method.

\section{IGConv for Arbitrary-Scale}
Since our method's core operators are convolution and depth-to-space, it is only able to upsample an integer scale.
While, by predicting $\lceil r\in \mathbb{R} \rceil$ and then bicubic downsampling to $r$, our methods can predict any arbitrary scales while this is not an optimal approach.
To validate the performance of this simple implementation, we train and evaluate IGConv$^{+}_{arb}$, which learns arbitrary float scales $r \in [1, 4]$ using the aforementioned methods to learn arbitrary-scale and compare it with ASSR methods on non-integer scales.
In this case, size factor $s_{r}$ can be estimated by arbitrary float scale $r$ while coordinates $C_{r}$ are still estimated from $\lceil r \rceil$.
As shown in Table~\ref{tab:sota_float}, despite its naive implementation, IGConv$^{+}_{arb}$ achieves results comparable to other methods.

\section{Visual Results on OOD Scale}
\begin{figure}[h!]
  \centering
  \includegraphics[width=0.8\columnwidth]{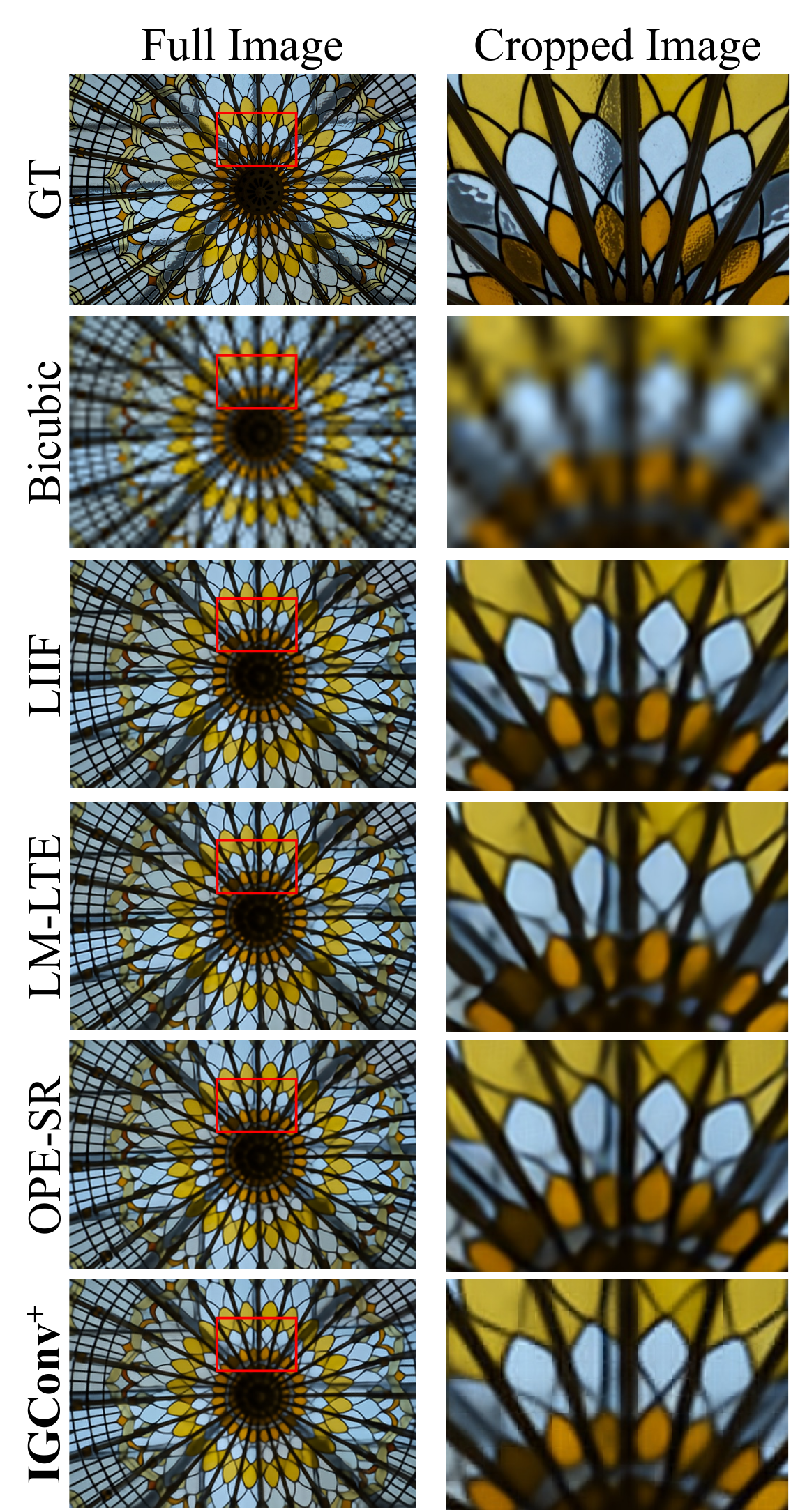}
  \caption{
    Visual comparisons on bicubic, ASSR methods~\cite{LIIF, LMF, OPESR}, and our IGConv$^{+}$ using RDN~\cite{RDN} encoder on out-of-distribution scale~($\times$24). Cropped images correspond to the red bounding boxed area of full images.
  }
  \label{fig:visual_ood}
\end{figure}

Since our main goal is training multiple integer scales simultaneously, IGConv$^{+}$ fails to predict OOD scale~($\times$24) reliably.
In the figure~\ref{fig:visual_ood}, IGConv$^{+}$ results in unpleasant artifacts in the areas where the color changes abruptly. 
However, note that our IGConv$^{+}$ successfully restores fine details such as thin lines, with significantly reduced inference latency about $\times$25 compared to LM-LTE~\cite{LMF}.
Thus, addressing these limitations while maintaining computational efficiency and performance should be considered as a future research work.

\end{document}